\documentclass[letterpaper]{article} % DO NOT CHANGE THIS
\usepackage{aaai25}  % DO NOT CHANGE THIS
\usepackage{times}  % DO NOT CHANGE THIS
\usepackage{helvet}  % DO NOT CHANGE THIS
\usepackage{courier}  % DO NOT CHANGE THIS
\usepackage[hyphens]{url}  % DO NOT CHANGE THIS
\usepackage{graphicx} % DO NOT CHANGE THIS
\usepackage{natbib}  % DO NOT CHANGE THIS AND DO NOT ADD ANY OPTIONS TO IT
\usepackage{caption} % DO NOT CHANGE THIS AND DO NOT ADD ANY OPTIONS TO IT
\usepackage{algorithm}
\usepackage{algorithmic}

\usepackage[utf8]{inputenc} % allow utf-8 input
\usepackage[T1]{fontenc}    % use 8-bit T1 fonts
\usepackage{url}            % simple URL typesetting
\usepackage{booktabs}       % professional-quality tables
\usepackage{amsfonts}       % blackboard math symbols
\usepackage{nicefrac}       % compact symbols for 1/2, etc.
\usepackage{microtype}      % microtypography
\usepackage{xcolor}         % colors

\usepackage{times}
\usepackage{soul}
\usepackage{url}
\usepackage[utf8]{inputenc}
\usepackage{graphicx}
\usepackage{amsmath}
\usepackage{amsthm}
\usepackage{booktabs}
\usepackage{algorithm}
\usepackage{algorithmic}
\usepackage{verbatim}
\usepackage{graphicx}

\usepackage{newfloat}
\usepackage{listings}
\DeclareCaptionStyle{ruled}{labelfont=normalfont,labelsep=colon,strut=off} % DO NOT CHANGE THIS
\floatstyle{ruled}
\newfloat{listing}{tb}{lst}{}
\floatname{listing}{Listing}
\nocopyright
\title{Motion-Zero: A Zero-Shot Trajectory Control Framework of Moving Object\\ for Diffusion-Based Video Generation}
\author{
    Changgu Chen\textsuperscript{\rm 1}
    Junwei Shu\textsuperscript{\rm 1},
    Gaoqi He\textsuperscript{\rm 1}, 
    Changbo Wang\textsuperscript{\rm 2}\equalcontrib, 
    Yang Li\textsuperscript{\rm 1}\equalcontrib,
}
\affiliations{
    \textsuperscript{\rm 1}School of Computer Science and Technology, East China Normal University, Shanghai, China \\
    \textsuperscript{\rm 2}School of Data Science and Engineering, East China Normal University, Shanghai, China \\
    \{52215901006, 51265902091\}@stu.ecnu.edu.cn, \{gqhe, cbwang, yli\}@cs.ecnu.edu.cn, 
}

\usepackage{bibentry}
\begin{document}

\maketitle

\begin{abstract}
Recent large-scale pre-trained diffusion models have demonstrated a powerful generative ability to produce high-quality videos from detailed text descriptions. 
However, exerting control over the motion of objects in videos generated by any video diffusion model remains
a challenging problem.
In this paper, we propose a novel zero-shot moving object trajectory control framework, Motion-Zero, to enable arbitrary single-object-trajectory control for the text-to-video diffusion model.  
To this end, an initial noise prior module is designed to provide a position-based prior to improve the stability of the appearance of the moving object and the accuracy of position.
In addition, based on the attention map of the U-Net, spatial constraints are directly applied to the denoising process of diffusion models, which further ensures the positional consistency of moving objects during the inference.
Furthermore, temporal consistency is guaranteed with a proposed shift temporal attention mechanism.
Our method can be flexibly applied to various state-of-the-art video diffusion models without any training process.
Extensive experiments demonstrate our proposed method can control the motion trajectories of arbitrary objects while preserving the original ability to generate high-quality videos.
\end{abstract}


\section{Introduction}

In recent years, the generative capabilities of diffusion models have been widely recognized in both text-to-image~\cite{dhariwal2021diffusion,ho2020denoising,rombach2022high,song2020denoising} and text-to-video domains~\cite{guo2023animatediff,ho2022imagen}.
Although these video models are capable of producing high-definition, high-resolution, and fluid video animations,
the dynamic motion trajectories of the generated objects are relatively random in existing text-to-video models~\cite{blattmann2023stable,zhang2023show}.
Thus, accurate control of an object's motion trajectories in a generated video remains rudimentary. 

\begin{figure*}[t]
  \centering
   \includegraphics[width=0.8\linewidth]{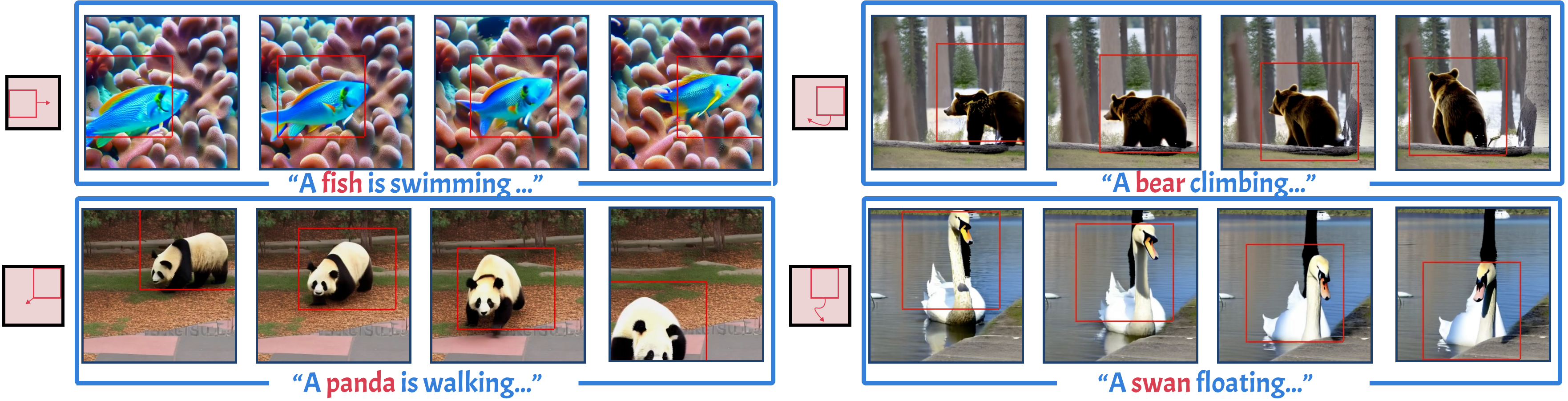}

   \caption{Our Motion-Zero framework endows different pre-trained video diffusion models with the capability to manipulate object trajectories directly, circumventing the need for supplementary training. By designating the target entity in the input prompts and a sequence of bounding boxes, users can intuitively direct the motion path of the object within the generated video.} 
   \label{fig:fig1}
   \vspace{-3mm}
\end{figure*}

Various strategies have been proposed to address this problem.
Some methods~\cite{Zhang2023AddingCC,wu2023tune} try to control the trajectory of the moving object by providing a detailed conditional motion sequence, such as a dancing skeleton. 
However, the cost of acquiring conditional control sequences is non-negligible and inhibits the diverse generation of video outputs by the user.
% leads to reduced diversity in the generated content
To obtain variety over the generated video with more control capability, i.e. trajectories of moving objects, several methods~\cite{wang2023motionctrl,wang2023videocomposer,yin2023dragnuwa} harness a substantial dataset of motion and trajectory pairs to train the baseline models.
Although the results are very impressive and promising, these methods are solely applicable to the base models on which they were trained and cannot be applied to other models directly.
Furthermore, all of these methods require extensive training and significant computational resources, preventing ordinary users from using them.

Nevertheless, the fundamental text-to-video diffusion models \cite{blattmann2023stable,zhang2023show} have undergone an extremely large scale training dataset~\cite{Bain21,xue2022advancing}. 
Theoretically, the pre-trained model should inherently have learned extensive knowledge about the dynamics of a variety of object movements.
However, the semantics of the latent space in different modules of a video diffusion model are not explicitly defined, which makes it difficult to manually control generated video in desired motion dynamics.
To some extent, the methods mentioned above try to align and define those latent spaces in the baseline model by training with additional labeled datasets, and successfully harnessing the intrinsic knowledge of pre-trained models to control the motion.
In addition, we observe that the initial noise plays a significant inspirational role in the generation of the videos.
The same initial noise tends to produce videos with similar content.
With these two characteristics, we can manipulate the trajectory of moving objects.

In this paper, we propose a zero-shot trajectory control framework, Motion-Zero, to guide the motion of generated objects in video. Our proposed method can be easily applied to pre-trained video diffusion model while achieving universality and plug-and-play capabilities of controlling. 
To this end, we first design an Initial Noise Prior Module to provide noises based on the motion trajectories given by the user. Then, inspired by the fact that the values of the cross-attention map within the U-Net largely determine the generation location of the subject mentioned in the prompt~\cite{xie2023boxdiff,epstein2023diffusion}, 
Spatial Constraints with an attention similarity loss over these attention maps are proposed to achieve precise manipulation of object positioning within individual frames.
Furthermore, we observe that merely imposing spatial constraints on the positions of objects can negatively impact the overall coherence and video quality.
To preserve the continuity throughout the video sequence, a novel Shift Temporal Attention Mechanism is proposed to further maintain the baseline model focus on the same objects across the time axis. 
Through the implementation of these modules, our proposed framework can effectively generate high-quality video controlled by bounding-box trajectories without any training process. 
Samples generated by Motion-Zero can be viewed in Fig.\ref{fig:fig1}.
The main innovations of our work can be summarized as follows,
\begin{itemize}
\item We propose a zero-shot framework Motion-Zero, which is capable of controlling 
generated object for arbitrary trajectory within a pre-trained video generation diffusion model. Our Motion-Zero is plug-and-play and without any additional training. 

\item Our Initial Noise Prior Module is a novel way to acquire semantic initial noise for high-quality generation. Moreover, Spatial Constraints and Shift Temporal Attention Mechanism respectively exploit the cross-attention and temporal-attention dimensions to obtain spatial control and temporal consistency.

\item Extensive experiments demonstrate that our proposed work can be applied to pre-trained video diffusion models. Our method enables the original baseline to generate objects with arbitrary trajectories.
\end{itemize}

\section{Related Works}
With the rapid development of deep learning \cite{NEURIPS2023_a24cd16b,DBLP:conf/eccv/Zheng-bttdm-eccv, DBLP:conf/mm/Zheng-ddpmbs-mm, wang2023rethinking, sun2024rethinking, sun2023contrastive, li2024groundinggpt, li2024unifiedmllm, wang2024advancing, wang2024qcrd, ijcai2024p762,cai2021leveraging,cai2022global,cai2023explicit, chen2024multi,chen2023video, 9720180,9966829,10.1145/3544548.3581067,10094642,sun2024chattracker}, 
there are pre-trained methodologies in place that have the capability to control the motion trajectories of objects within generated videos. 
VideoComposer \cite{wang2023videocomposer} employs a two-stage training strategy to incrementally incorporate temporal information and control signals. 
A motion condition encoder is proposed for training, designed to enhance the model's ability to understand and integrate motion-specific information.
DragNUMA \cite{yin2023dragnuwa} also employs a comparable training strategy and utilizes optical flow as a conditioning mechanism for trajectory modeling.
MotionCtrl \cite{wang2023motionctrl} synthesizes the approaches of these two works and introduces a camera control module and a trajectory control module to further refine the management of movement within generated video content.
Boximator \cite{wang2024boximator} proposes a novel self-tracking technique to simplify the learning of box-object relationships.
MotionBooth \cite{wu2024motionbooth} allows for controlling the movement of a specific object by fine-tuning a photo of that object.
\cite{yu2024zero} enables the movement of a specified object in a user's input photo to generate a video. 
Motion-I2V \cite{shi2024motion} leverages a two-stage training process, allowing users to alter the motion in the generated video by dragging.
These models typically rely on extensive training on large-scale datasets such as WebVid-10M \cite{Bain21} and HD-VILA-100M \cite{xue2022advancing} which lead to expensive training costs.
Also, these models are typically constrained to operate on the specific models they were trained on, lacking the flexibility to interchange base video diffusion models. 
This limitation hinders their adaptability and limits the scope of their application to only those scenarios for which they were explicitly designed.
Our proposed model stands out in that it can be applied to any base diffusion model without the necessity for further training. 
% This versatility not only broadens its applicability but also results in substantial savings in training costs and resources, offering a significant advantage over more rigid, model-specific methods.
Recently, Trailblazer~\cite{Ma2023TrailBlazerTC} enhances attention on the box area in cross-attention with a zero-shot setting, similar to our work.
In contrast, our operation on cross-attention is supervised by losses, providing better interpretability and effectiveness.
% 加入peekaboo
Peekaboo~\cite{jain2024peekaboo} uses a masked attention module to achieve control without the need for training.
FreeTraj~\cite{qiu2024freetraj} utilizes frequency fusion to generate results whose trajectory is aligned with the given box.
Although these methods exploit zero-shot settings in controllable video generalization, 
our proposed method emphasises on the importance of the initial noise and the control of spatial and temporal consistency, and achieves superior performance.

\section{Preliminaries}
\textbf{Video Diffusion Model:}
Video diffusion models are designed to produce high-quality and diverse videos, guided by text prompts. 
% However, generating videos at the pixel level is computationally intensive. To address this, latent video diffusion models serve as an efficient alternative. 
To save the computational costs, \cite{rombach2022high} utilize a U-Net as a denoising model within a latent space, significantly reducing the computational load in terms of both time and space.

In detail, these models employ a Variational Autoencoder (VAE), which comprises an encoder $\mathcal{E}$ and a decoder $\mathcal{D}$. The encoder compresses the original video from pixel space into a latent representation, and the decoder reconstructs the video from this latent space back to pixel space. The 3D U-Net typically consists of a series of down-sampling blocks, middle blocks, and up-sampling blocks. Each block is equipped with convolutional layers, spatial transformers, and temporal transformers. The optimization of the 3D U-Net (denoted as $\epsilon_{\theta}$) is executed through a noise prediction loss function:
\begin{equation}
    \mathcal{L} = \mathcal{E}_{\mathbf{z}_0,c,\epsilon\sim N(0,I),t}[||\epsilon-\epsilon_{\theta}(\mathbf{z}_t,t,\mathbf{c})||^2_2],
\end{equation}
where $z_0$ is the latent code of the training videos, c is the text prompt condition, $\epsilon$ is the Gaussian noise added to the latent code, and $t$ is the time step. The noised latent code $\mathbf{z}_t$ is determined as:
 \begin{equation}
    \mathbf{z}_t = \sqrt{\overline{a}_t}\mathbf{z}_0+\sqrt{1-\overline{a}_t}\mathbf{\epsilon},\overline{a}_t = \prod^t_{i=1}a_t,
    \label{eq_z_denoise}
\end{equation}
where $a_t$ is a hyper-parameter used for controlling the noise strength based on time $t$.

\noindent\textbf{Motion Trajectory Control:}
\label{setting}
%% Motion Trajectory Control is used to control 精细化的物体运动轨迹生成 \cite{}
%% 介绍怎么控制
% a straightforward idea is to exert some control over the trajectory of object movement in the generated videos.
% However, existing trajectory control methods either require large-scale video training data or rely on a template video, leading to a low degree of freedom in video generation capabilities.
% To enable the algorithm to control any video diffusion model without the need for training, we propose a new setting: zero-shot motion trajectory control.
Based on video diffusion, the task of motion trajectory control is to precisely control the motion trajectory of objects in generated videos.
The optimization objective can be formulated as:
\begin{equation}
    \mathcal{L} = \mathcal{E}_{\mathbf{z}_0,c,\epsilon\sim N(0,I),t}[||\epsilon-\epsilon_{\theta}(\mathbf{z}_t,t,\mathbf{c},\mathcal{B})||^2_2].
    \label{eq:mot}
\end{equation}
Specifically, users can input a text condition $\mathbf{c}$ along with a sequence of rectangular boxes $\mathcal{B}=\{(x_1^f,y_1^f),(x_2^f,y_2^f)\}^{N_f}$, 
where $(x_1^f,y_1^f),(x_2^f,y_2^f)$ are the upper left and lower right points of the box in frame $f$, $N_f$ is the total frame number. The boxes $\mathcal{B}$ correspond to the position of the moving object.

% The model is then capable of generating a video in which the object moves within the specified locations of these rectangular boxes without any training.
% Based on this setting, our Motion-Zero framework operates entirely during the inference stage, thus eliminating the need for any training and making it compatible with any pre-trained video diffusion model.

\begin{figure*}[t]
  \centering
   \includegraphics[width=0.75\linewidth]{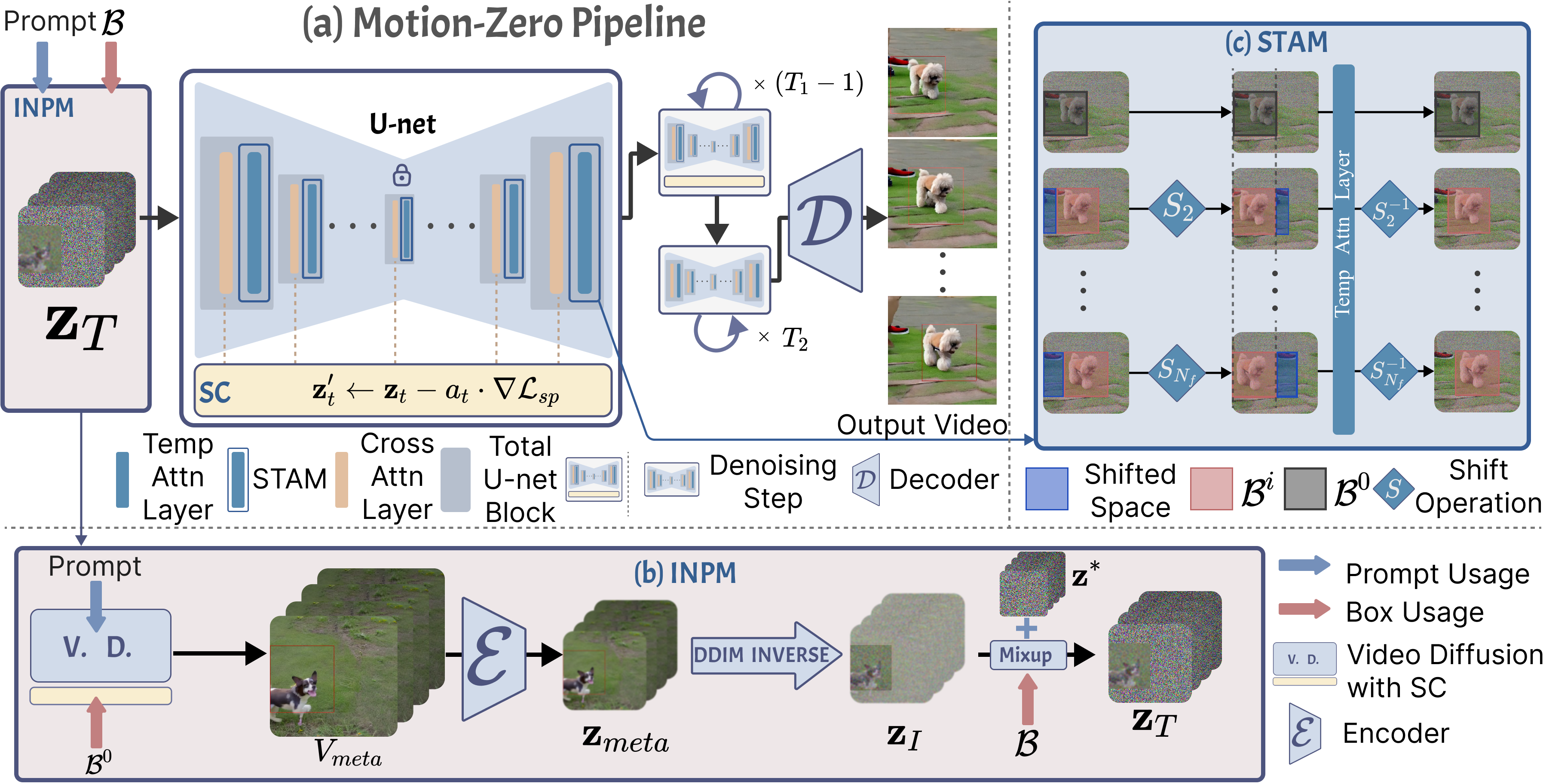}

   \caption{Overview of our Motion-Zero. The total pipeline is shown in (a). Given the box condition $\mathcal{B}$ and the prompt condition, we generate the prior latents $\mathbf{z}_T$ by our Initial Noise Prior Module (INPM) as shown on (b). At timestep $t$, 
 $\mathbf{z}_t$ is firstly optimized to $\mathbf{z}_t'$ by the Spatial Constraints (SC). Subsequently, $\mathbf{z}_t'$ is passed to the UNet with Shift Temporal Attention Module (STAM) as demonstrated on (c). All the parameters of the video diffusion are frozen. $T_1$ represents the number of timesteps during which SC and STAM are applied, and $T_2$ denotes the number of timesteps where the original video diffusion process is utilized.} 
   \label{fig:fig2}
   \vspace{-3mm}
\end{figure*}

\section{Methodology}
\subsection{Overview}

% Our task is crafted to operate without the necessity for training. 
%% motion zero 先用什么什么。。。（Section 4.2） 再利用用什么实现的 （Seciton 4.3） 
% It allows users of any pre-trained video diffusion model to input a textual prompt alongside a series of rectangular boxes. 
% These boxes correspond to the position of an object referenced within the control prompt. 
% The model is adept at producing videos where the object is seen moving within the bounds of these rectangles. 

%%之前的轨迹控制需要training , 因此我们提出了 zero-shot framework. 它通过三个模块来实现zero-shot，哪三个模块。具体来说，
Existing trajectory control methods require large-scale video training data to optimize Eq.\ref{eq:mot}, leading to a high computational cost.
% low degree of freedom in video generation capabilities. 
% To address this issue, a new setting is proposed: 
Differently, our work falls into the setting of zero-shot motion trajectory control.
To this end, our proposed Motion-Zero framework operates entirely during the inference stage, thus eliminating the need for training and making it compatible with various pre-trained video diffusion models.
The pipeline is shown in Fig.\ref{fig:fig2} (a). 
In the following section, we provide a detailed presentation of our proposed Motion-Zero framework and its components.
Firstly, we introduce a noise prior generation module to produce the initial noise for diffusion. Then, we describe the designed constraints and loss of the moving object position and spatial consistency. Finally, our shift temporal attention mechanism is presented to further improve the temporal consistency. 

\subsection{Initial Noise Prior Module}
\label{init}
% Building on some prior work \cite{Wu2023FreeInitBI}, we have observed that 
According to the theory of DDIM Inversion \cite{ho2020denoising}, the initial noise has a significant impact on the final generated outcome. 
We introduce Initial Noise Prior Module (INPM) to leverage this property to provide a strong prior for the position of the moving object. 
Several steps are involved to integrate a moving object into a sequence of frames with a coherent prior as shown in Fig.\ref{fig:fig2} (b). 
Firstly, given a prompt $\mathbf{c}$ and the boxes in the same location with $\{\mathcal{B}^0\}^{N_f}$, a meta video $V_{meta}$ is sampled,  $\mathbf{z}^*\sim\mathcal{N}(0,\mathbf{I}) $ as latent input, using the baseline video diffusion model and our proposed spatial constraints (introduced in the next section). 
This generated video has the target object staying at the location with $\{\mathcal{B}^0\}^{N_f}$ due to the spatial constraints.
It is noteworthy that controlling the model to generate standing-still objects is much easier than controlling the movement of objects.
Then, a video latent $\mathbf{z}_{meta}$ is generated based on $V_{meta}$ from Encoder $\mathcal{E}$.
Once $\mathbf{z}_{meta}$ is prepared, we perform a DDIM Inversion to obtain the corresponding noise latent representation $\mathbf{z}_I$.  
We crop the latent representation within the box $\mathcal{B}^0$ for each frame, creating a sequence of latent patches containing the visual target. 
Subsequently, we use a local mixup operation \cite{zhang2017mixup} to mix the latent patches and the initial noise $\mathbf{z}^*$ in the range of $\mathcal{B}^f$ frame by frame.
Our INPM allows us to set a coherent prior in the corresponding object's position in the initial noises.
% by utilizing the influence of the  on the subsequent generation process. 
It also ensures that the animated object maintains consistency in appearance and movement across the video frames, without incurring additional computational costs during the generation process.
Details are shown in Alg.\ref{alg:init}.

% \begin{figure}[t]
%   \centering
%    \includegraphics[width=1\linewidth]{imgs/fig3.pdf}
%    %\includegraphics[width=0.8\linewidth]{egfigure.eps}

%    \caption{The details of the proposed SC. $\mathcal{L}_i$ and $\mathcal{L}_o$ are used to enhance the attention value inside $\mathcal{B}$ and suppress the attention value outside. $\mathcal{L}_c$ is to ensure the centroid of the attention value close to the center of $\mathcal{B}$. $\mathcal{L}_s$ enables the attention value inside $\mathcal{B}$ of different frames more similar.}
%    \label{fig:fig3}
% \end{figure}

\begin{algorithm}[tb]
    \caption{Initial Noise Prior Module}
    \label{alg:init}
    \textbf{Input}: $\mathbf{c}$, $\mathcal{B}$ \\
    \textbf{Parameter}: $N_f$, $\lambda_p$ \\
    \textbf{Function}:  $\mathbf{Pix}(\mathbf{a},\mathbf{b})$ means the elements get from tensor $\mathbf{a}$ in the range of box $\mathbf{b}$ \\
    \textbf{Output}: Initial noise $\mathbf{z}_T$
    \begin{algorithmic}[1] %[1] enables line numbers
        \STATE $\mathbf{z}^* \sim \mathcal{N}(0,\mathbf{I})$ \quad \textcolor{gray}{$\triangleright$ random sample the first latent code}
        \STATE $\mathbf{V}_{meta} \leftarrow \mathbf{VideoDiffusion}(\textbf{z}^*, \textbf{c}, \mathcal{B}_0) $ \quad \textcolor{gray}{$\triangleright$ Video Diffusion using SC, the first box and \textbf{c} as condition}
        \STATE $ \mathbf{z}_{meta} \leftarrow \mathbf{Encoder}(\mathbf{V}_{meta})$
        \STATE $\mathbf{z}_I \leftarrow \mathbf{DDIM\_Inverse}(\textbf{z}_{meta})$ \quad \textcolor{gray}{$\triangleright$ inverse the $\mathbf{z}_{meta}$}
        \STATE $\textbf{z}_T$ $\leftarrow$  $\textbf{z}^*$
        \FOR{all f=1,2,...,$N_f$}
            \STATE $\mathbf{Pix}(\mathbf{z}_T^f, \mathcal{B}^f)$$\leftarrow$$\lambda_p$$\cdot \mathbf{Pix}(\mathbf{z}^f_I, \mathbf{\mathcal{B}^0})$+ \\ $(1$-$\lambda_p)$$\cdot$$ \mathbf{Pix}(\mathbf{z}_T^f, \mathcal{B}^f)$  \quad \textcolor{gray}{$\triangleright$ local mixup operation}
        \ENDFOR
        \STATE return $\mathbf{z}_T$ \quad \textcolor{gray}{$\triangleright$ initial latent with position prior}
    \end{algorithmic}
\end{algorithm}

\subsection{Spatial Constraints with Attention Map}
\label{Spatial}
% The purpose of the spatial constraint is to confine the object's movement to specified coordinates while maintaining its spatial consistency.
The INPM alone is insufficient for precise manipulation of an object's trajectory. 
To further improve the capability of control, we introduce Spatial Constraints (SC) deployed at each denoising step $t$ to optimize the intermediate latent representation $\mathbf{z}_t$. 
This optimization is crucial for enhancing the accuracy of the moving object position and preserving spatial consistency.
Within the conditional denoising architecture \cite{rombach2022high}, cross-attention serves as the pivotal bridge that connects the text prompt with the content generated.
During the denoising steps, conditioned on the prompt $\mathbf{c}$ and the intermediate features $\mathbf{z}_t$, the corresponding cross-attention map can be obtained as $\mathbf{A}$:
\begin{equation}
    \mathbf{A} = \mathbf{Softmax}(\mathbf{QK}^\top/\sqrt{d}),
\end{equation}
\begin{equation}
    \mathbf{Q} = \mathbf{W}_Q \mathbf{z}_t, \mathbf{K} = \mathbf{W}_K \mathbf{c},
\end{equation}
where $\mathbf{Q},\mathbf{K}$ are the query and key with the $\mathbf{z}_t$ and $\mathbf{c}$, respectively. $\mathbf{W}_Q, \mathbf{W}_K$ are two learnable matrices, which are frozen in our settings. 
Assuming a maximum number $N_p$ of prompt tokens $\{\mathbf{p}_1,...,\mathbf{p}_{N_p}\}$ in condition $\mathbf{c}$, at time step $t$, it is feasible to derive $N_p$ cross-attention maps $\{\mathbf{A}_1,...,\mathbf{A}_{N_p}\}$. 

When a user specifies the index $k$ of prompt tokens $\mathbf{p}_k$ intended to control the trajectory of an object, along with the box $\mathcal{B}$, the box-loss can be employed to ensure that our model controls the object to appear within the input box in every frame.
Specifically, the box-loss is achieved through an optimization approach that maximizes the values of the $\mathbf{A}_k$ for the corresponding prompt tokens inside the box, while minimizing the values outside the box. 
Frame by frame, we scale down the user-specified coordinate box to the corresponding coordinates in the latent space and construct a mask $\mathbf{M}^f$ where $1$ for areas in the box and $0$ otherwise. 
To ensure the values inside the box for attention are maximized, we propose an intuitive solution ${\mathcal{L}_{i}}$ as
\begin{equation}
    {\mathcal{L}_{i}}^f = 1-\frac{1}{P}\sum \mathbf{g}(\mathbf{A}_k^f\cdot \mathbf{M}^f,P) ,
\end{equation}
% \begin{equation}
%     {\mathcal{L}_{innerbox}}_i = \sum^f {\mathcal{L}_{innerbox}}^f_i
% \end{equation}
where $\mathbf{g}(\cdot, P)$ means the top $P$ highest value will be selected. 
If we employ all the values within the attention maps, it could disrupt the stability of the denoising process. 
Conversely, utilizing too few values could result in a less pronounced control effect. 
% In this context, we adopt $P=0.8$ as an empirical threshold.
On the other hand, we aim for the attention values outside the box to be as minimal as possible which is formulated as 
% ${\mathcal{L}_{o}}^f$:
\begin{equation}
    {\mathcal{L}_{o}}^f = \frac{1}{P}\sum \mathbf{g}(\mathbf{A}_k^f\cdot (1-\mathbf{M}^f),P).
\end{equation}
% \begin{equation}
%     {\mathcal{L}_{outerbox}}_i = \sum^f {\mathcal{L}_{innerbox}}^f_i
% \end{equation}
Within these two constraints, we can ensure that the object generated in each frame is contained within the box. 
However, there is no guarantee that the object will be positioned at the center of the box. 
To mitigate this, we propose another center-loss $\mathcal{L}_{c}$ to encourage the centroid of the $\mathbf{A}_k^f$ to closely align with the center of the box as%
\begin{equation}
\resizebox{0.9\hsize}{!}{$(W_{\mathbf{A}_k^f}, H_{\mathbf{A}_k^f}) = \frac{1}{\sum_{w,h}\mathbf{A}_{k,w,h}^f}(\sum_{w,h}w\cdot\mathbf{A}_{k,w,h}^f,\sum_{w,h}h\cdot\mathbf{A}_{k,w,h}^f)$,}
\end{equation}%
\begin{equation}
            {\mathcal{L}_{c}}^f = ||(\frac{x_1^f+x_2^f}{2},\frac{y_1^f+y_2^f}{2}) - (W_{\mathbf{A}_k^f}, H_{\mathbf{A}_k^f})||_1,
\end{equation}
where $\mathbf{A}_{k,w,h}^f$ is the value of $\mathbf{A}_{k}^f$ at the position of $(w,h)$, $(W_{\mathbf{A}_k^f}, H_{\mathbf{A}_k^f})$ is the centroid position of $\mathbf{A}_k^f$ with the $k$-th token  concept. 
After establishing robust control over the position of the object, we recognize unexpected variations in the object's appearance due to the substantial extent of movement.
To ensure appearance consistency across frames, we strive to maintain the uniformity of the $\mathbf{A}_k^f$ within the box. To this end, we introduce a similarity loss $\mathcal{L}_{s}$ as
\begin{equation}
\resizebox{0.85\hsize}{!}{$
            {\mathcal{L}_{s}} = 1-\frac{1}{N_f-1}\sum_{f = 1}^{N_f-1}\mathbf{Sim}(\mathbf{Pix}(\mathbf{A}_i^f, \mathcal{B}^f), \mathbf{Pix}(\mathbf{A}_i^{f+1}, \mathcal{B}^{f+1}))$},
\end{equation}
where $\mathbf{Sim}(\cdot,\cdot)$ means the similarity of two elements and $\mathbf{Pix}()$ is a function getting corresponding elements of the map $\mathbf{A}^f$ within a box range $\mathcal{B}^f$. Cosine similarity is adopted here. 
At each timestep, the overall spatial constraints $\mathcal{L}_{sp}$ are formulated as follows:
\begin{equation}
            {\mathcal{L}_{sp}} = \sum_{f}(\lambda_i{{\mathcal{L}_{i}}^f+\lambda_o\mathcal{L}_{o}}^f+\lambda_c{\mathcal{L}_{c}}^f) + \lambda_s{\mathcal{L}_{s}},
\label{eq_total}
\end{equation}
where $\lambda_i,\lambda_o,\lambda_c,\lambda_s$ are hyper-parameters. By minimizing and calculating the gradient of the Eq.\ref{eq_total}, we can optimize our latent $\mathbf{z}_t$ in Eq.\ref{eq_z_denoise} as follows: 
\begin{equation}
    \mathbf{z}_t' \leftarrow \mathbf{z}_t - \beta_t \cdot \nabla {\mathcal{L}_{sp}},
    \label{eq:update}
 \end{equation}
where $\beta_t$ linearly decays at each timestep $t$.
Specifically, before denoising with the U-Net at each timestep $t$, we update $\mathbf{z}_t$ to $\mathbf{z}_t'$ using Eq.\ref{eq:update}. Then, we continue the denoising process using $\mathbf{z}_t'$.
Under the combined effect of the aforementioned constraints, the latent variable $\mathbf{z}_t$ at each timestep will gradually shift towards generating high-response attention in the specified position, while ensuring that the appearance attributes of the object within the box remain unchanged. Consequently, the target object is synthesized within the bounding box area provided by the user.

\subsection{Shift Temporal Attention Mechanism}
\label{temp}
After applying the spatial constraints on attention maps, pre-trained video diffusion models still encounter difficulties in generating sequences of continuous actions. Within the temporal module of the diffusion process, the latent representation is reshaped into the following configuration: %    \mathbf{z}' = 
\begin{equation}
 z_{(b \cdot H \cdot W + h \cdot W + w, f, c)}' = z_{(b, f, c, h, w)}
\end{equation}
where $z'$ represents the result of applying the rearrange operation to $z$.
Within the temporal transformer, attention is focused on the same pixel across different frames. 
This leads to a scenario where, if the extent of motion is too large, the same position in different frames could undergo significant semantic changes, resulting in a lack of coherence in the generated dynamics.
% A very straightforward solution to this problem is to expand the capabilities of the temporal transformer module. 
% This enhancement would enable it not only to perceive the same locations across different frames but also to recognize different positions across those frames. 
% This can be illustrated as follows (using einops notation):
% \begin{equation*}
%     \mathbf{z}' = rearrange(\mathbf{z}, b \ f \ c \ h \ w \rightarrow b \ (f\ h \ w) \ c),
% \end{equation*}
% Indeed, such an expansion of the temporal transformer module would lead to substantial computational resource consumption during both training and inference phases. 
% Moreover, it would require the model to be trained from the beginning, and this approach cannot be directly applied to currently pretrained video diffusion models. 
% This limitation exists because the expanded temporal transformer would alter the architecture in a way that is incompatible with the weights learned during the pretraining of the existing models.The purpose of the operations is

To overcome this inconsistency, we propose a Shift Temporal Attention Mechanism (STAM) to improve the dynamics of the moving object in different frames. 
Specifically, we shift the elements of $\mathbf{z}^f$ inside the $\mathcal{B}^f$ range with the elements inside the $\mathcal{B}^0$ range and use the overlapped parts to fill in the vacated spaces, as shown in Fig.\ref{fig:fig2} (c).
Therefore, the subsequent frames within the box range can be aligned with the box range of the first frame.
The steps are shown:
\begin{equation}
\begin{split}
   \mathbf{z}_w^f &= \mathbf{Shift}(\mathbf{z}^f, \mathcal{B}^f, \mathcal{B}^0) ,\\
    {\mathbf{z}_w}' &= \mathbf{TemporalAttention}(\mathbf{z}_w) ,\\
    {\mathbf{z}^f}'&= \mathbf{Shift}({\mathbf{z}_w^f}', \mathcal{B}^0, \mathcal{B}^f) ,\\ 
\end{split}
\end{equation}
where $\mathbf{Shift}(\cdot, a,b)$ is the shift operation, ${\mathbf{z}^f_w}$ means the shifted latent ${\mathbf{z}^f}$ of frame $f$ and ${\mathbf{z}_w} = [{\mathbf{z}^1_w},..,{\mathbf{z}^f_w}]$.%,   that swaps the matrix in box $a$ and $b$. 
By applying STAM to a $\mathbf{TemporalAttention}$ in the baseline video diffusion models, we can achieve coherence in the motion of moving objects without additional training, and without incurring extra computational costs during inference. 
Note that we do not use STAM in INPM as the motion of objects in the meta video $V_{meta}$ generated within INPM occurs at the same location. Therefore, there is no need to shift temporal attention to align the positions of moving objects.

\section{Experiments}

We evaluate the effectiveness of our method from both qualitative and quantitative perspectives. 
In our experiments, the default baseline is ZeroScope \cite{Sterling2023ZeroScope}. 
ModelScope \cite{wang2023modelscope} is also employed to show that 
our method can be applied to various video diffusion baselines.
% our approach maintains generative performance with various baselines.
% the SOTA pre-trained model MotionCtrl~\cite{wang2023motionctrl} and 
TrailBlazer~\cite{Ma2023TrailBlazerTC} and Peekaboo~\cite{jain2024peekaboo} are involved in experiments to demonstrate the control ability of our proposed method.
Following TrailBlazer, 33 prompts containing different moving objects and motion patterns are employed as the evaluation dataset. For simple motions, we employed eight movement trajectories. 
For experiments involving complex trajectories, we utilized 17 randomly generated motion curves. 
Please refer to the appendix for specific prompts and trajectory parameters.

\subsubsection{Implementation Details}
\label{imple}

Our algorithm is fully implemented during the inference stage, thus it does not require any training.
The hyper-parameter $\lambda_i,\lambda_o$ are set to 1, $\lambda_c$ is set to 0.05, $\lambda_s$ is set to 0.5, $\lambda_p$ is set to 0.8.
To balance the trade-off between the size of GPU memory consumption and the semantic information retained in the attention map $\mathbf{A}$, we choose a $48\times48$ size for $\mathbf{A}$ when the output resolution is $384\times384$.
We use DDIM \cite{song2020denoising} as our sampling method.
In the experiment, $T_1$ in Fig.\ref{fig:fig2} is set as 10, which means
the SC and the STAM are employed during the first 10 timesteps; $T_2$ is set to 20, thus the total denoising timestep $T$ is 30. 
All of our experiments are conducted on a single NVIDIA A100 GPU.

\subsection{Comparisons with SOTA Methods}
\subsubsection{Qualitative Analysis.}

\begin{figure}[t]
  \centering
  \includegraphics[width=0.8\linewidth]{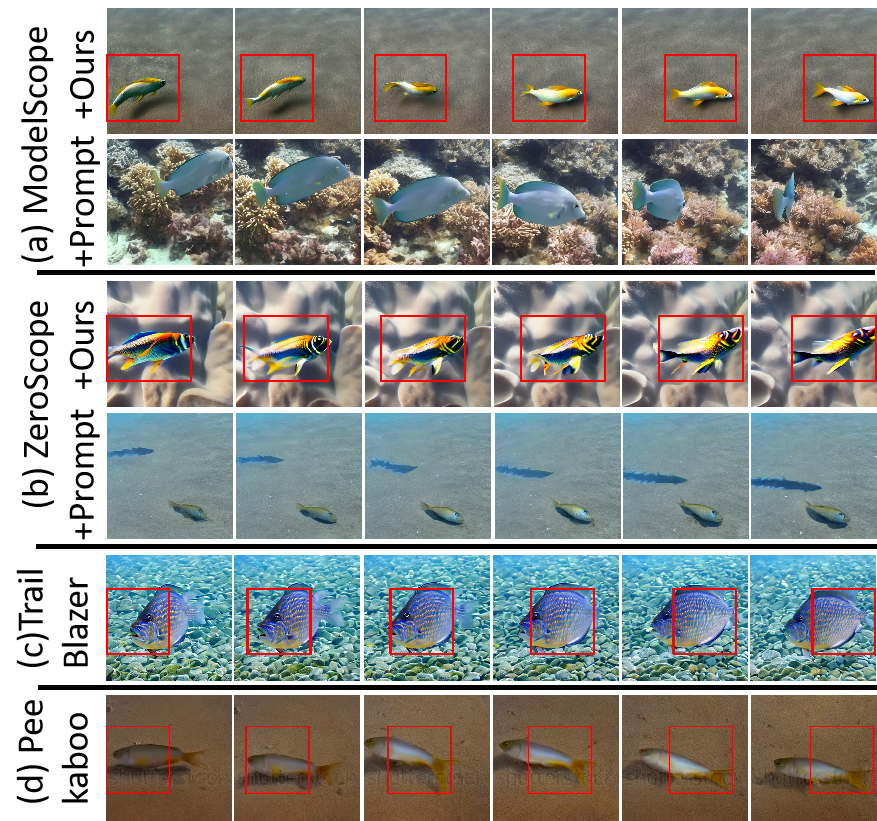}
   \caption{Quality comparison results on different methods. We take one frame from every three frames. The input prompt: \textit{A fish is swimming in the sea.} We employed ModelScope (a) and ZeroScope (b) as our baseline models and compared the effect of incorporating additional prompts with the integration of our Motion-Zero. In addition, we conducted a comparative analysis with TrailBlazer and Peekaboo.
   % pre-trained MotionCtrl model (c). 
   % The red dots serve as the control conditions. 
   }
   \label{fig:compare}
   \vspace{-4mm}
\end{figure}

\begin{figure*}[t]
  \centering
  \includegraphics[width=0.8\linewidth]{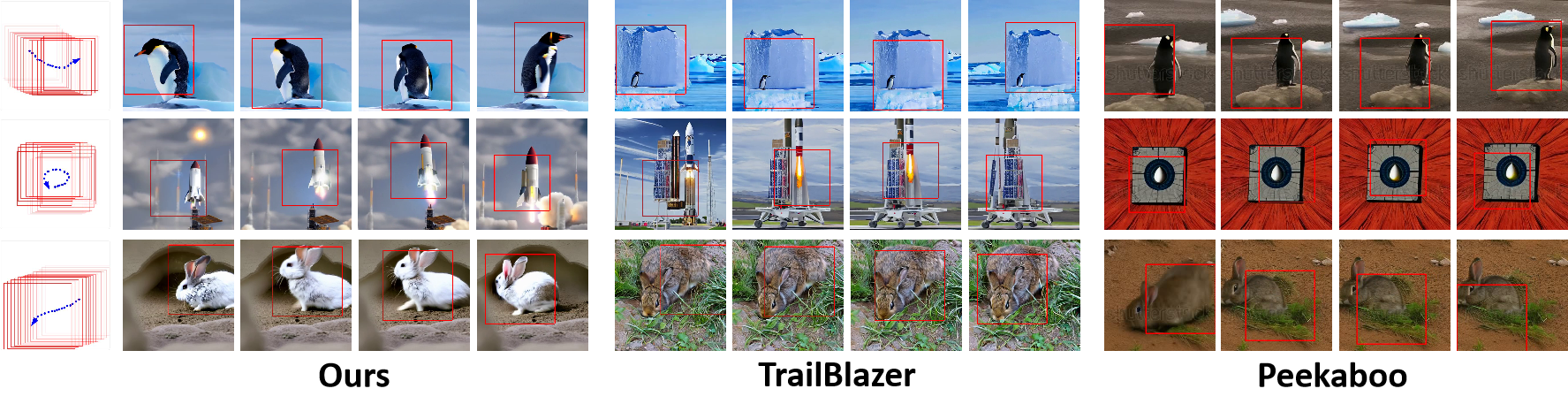}
   \caption{Quality comparison results with complex trajectories. We take one frame from every three frames. The input prompt of the first row: \textit{A penguin standing on an iceberg.} The second row: \textit{A rocket launching into space from a launchpad.} The third row: \textit{A rabbit burrowing downwards into its warren.}  \textbf{Zoom in for the best view.}}
   \label{fig:complex}
   \vspace{-3mm}
\end{figure*}

The results of the qualitative experiments with simple and complex trajectories are shown in Fig.\ref{fig:compare} and Fig.\ref{fig:complex}, respectively.
%we demonstrate qualitative experiments on simple trajectories. \LY{The base model depicted in the first two rows of the figure is ModelScope.} 
Our proposed Motion-Zero (+Ours) applied on other baselines (ModelScope and ZeroScope) can greatly increase the controllability of the objects' motion trajectories.
In Fig.\ref{fig:compare}, baselines (+Ours) refers to applying MotionZero to the baseline methods, while original baselines (+Prompt) have an extra prompt added to indicate the motion of objects: \textit{moving from left to right.}
As shown in Fig. \ref{fig:compare}, Motion-Zero correctly guides the motion of fishes by following the specific trajectories indicated by the red boxes. In contrast, 
both original ModelScope (+Prompt) and ZeroScope (+Prompt) fail to control the objects following the expected trajectories.
Fig.\ref{fig:compare} (c) demonstrates the generated results of TrailBlazer. 
It shows that the fish generated by TrailBlazer does not strictly follow the red bounding boxes. 
Fig.\ref{fig:compare} (d) demonstrates the results of Peekaboo. It is observed that the object detaches from the box in the last few frames, and the movement direction of the fish on the left does not align with the direction of the box.

Fig.\ref{fig:complex} indicates the comparison between ours, TrailBlazer, and Peekaboo in complex trajectories setting. Our method effectively controls the motion of objects for any trajectory, e.g. a flying back rocket. On the contrary, TrailBlazer exhibits cases where moving objects are coupled with the background motion, e.g. the penguin is relatively stationary. Furthermore, the rocket and the rabbit are losing control and the motion becomes irrelevant with the bounding boxes.
The penguins generated by Peekaboo lack dynamism, maintaining a single pose while only the scene moves. It fails to properly generate rockets, which might be due to the unusual nature of rotating and moving rockets. Regarding the rabbit scene, the content generated by Peekaboo exhibited frame skipping.

\subsubsection{Quantitative Analysis.}
% \noindent\textbf{Experiment Setting.}
Following LOVEU-TGVE competition \cite{wu2023cvpr}, we use the CLIP score \cite{hessel2021clipscore} to verify text-video consistency (Text Align) and inter-frame consistency (Consistency). The PickScore \cite{kirstain2023pick} is employed to predict user preferences of our model.
To further evaluate the control capability of our model, we employ metrics including mIoU, AP50, Cov., and CD. mIoU stands for mean Intersection-over-Union of the detected bounding boxes and the input box on the generated video. This metric is primarily used to assess whether the model can effectively control the position and size of moving objects. We use the OWLViT-large detector \cite{minderer2022simple} to detect object boxes in the generated video. AP50 refers to average precision@50\%, which is used to determine if the overlap between the detected boxes and the user-provided boxes is greater than 50\%. Cov. is used to assess the probability of the detector detecting moving objects. CD represents the distance between the center of the moving object in the generated video and the center of the user-provided box, reflecting whether the generated center of the moving object is consistent with the user input.

\begin{table}[t]
    \centering
    
    \resizebox{0.48\textwidth}{!}{%
    \begin{tabular}{cccc|cccc}
        \toprule
                  & Align & Cons. & Pick. & mIoU & AP50 & Cov. & CD($\downarrow$) \\
        \midrule
        ZeroScope   & 20.31          & 0.88      & 18.98 & - & - & - & - \\
        \quad +Ours & 21.96          & \textbf{0.94}        & \textbf{19.89}  & \textbf{0.54} & \textbf{0.64} & \textbf{0.96} & \textbf{0.07} \\
        ModelScope  & 20.55          & 0.91     & 18.70 & - & - & - & - \\
        \quad +Ours & \textbf{23.54}           & 0.90        & 19.56  & \textbf{0.54} & \textbf{0.64} & 0.88 & 0.08 \\
%        MotionCtrl  & 20.94          & 0.92      & 19.62 \\
         TrailBlazer & 21.53         & 0.92       & 19.64 & 0.52 & \textbf{0.64} & 0.91 & 0.14 \\ 
         Peekaboo & 20.04         & 0.83       & 18.70 & 0.43 & 0.45 & 0.89 & 0.11 \\
        \midrule
        \midrule
%        MotionCtrl (comp.)   & 20.94          & 0.92      & 19.62 \\
         TrailBlazer (c.) & \textbf{23.34}         & 0.92       & 18.81 & 0.50 & 0.61 & 0.91 & 0.13 \\
         Peekaboo (c.) & 20.89         & 0.84       & 18.53 & 0.41 & 0.42 & 0.88 & 0.11 \\
        Ours (c.) & 22.40         & \textbf{0.95}       & \textbf{19.19} & \textbf{0.55} & \textbf{0.67} & \textbf{0.97} & \textbf{0.07} \\
        \bottomrule
    \end{tabular}
    }
    \caption{Automatic metric on baseline methods and SOTA methods. The left half indicates the quality of the generation, while the right half demonstrates the control capability of the model. All metrics expect CD to be such that higher values($\uparrow$) indicate better performance. (c.) means the method is tested in complex trajectories. Align means Text Align, Cons. means Consistency and Pick. means PickScore.}
    \label{tab:metric}
\end{table}

% \noindent\textbf{Quantitative Results.}
As shown in Tab.\ref{tab:metric}, the left part indicates the generation quality, and it shows that our proposed Motion-Zero framework does not compromise the baseline model's performance but increases video generation quality instead. This demonstrates that our method exploits and preserves the model's generation capability even with the zero-shot setting.
In addition, the motion subject constrained by the box results in improved semantic accuracy and temporal consistency.
This indicates that our proposed Motion-Zero can get a better generation quality score due to the effectiveness of consistency between the prompt and motion.

% Upon examining the fifth row of the table, it is evident that Motion-Zero achieves an enhancement in performance over pre-trained methodology.
% the motion trajectory of objects does not 
% while the right component demonstrates the control capability. 
For the left in Tab.\ref{tab:metric}, our method outperforms Trailblazer in terms of control performance for both simple and complex trajectories and surpasses Peekaboo in every aspect. 
Trailblazer performs better on the Align metric because it manually modifies cross-attention rather than deriving it through loss. However, this diminishes its performance on other metrics.
Through mIoU and AP50 metrics, we can see that the size and position of the moving objects generated by our method have a closer match for the boxes provided by the users. This confirms that $L_o$ and $L_i$ indeed ensure that the generated objects remain within the boxes given by the users. Higher Cov. score demonstrates that the objects generated by our method are clearer. This proves that our INPM and STAM modules enable smoother motion of objects, while $L_s$ maintains the consistency of the objects. At the same time, CD indicates that the centers of the objects generated by our method are closer to the centers of the boxes provided. This reflects that $L_c$ can position the generated objects at the center of the box.

% \subsubsection{User Study \LY{might merge to previous section}} 
% \textcolor{blue}{done}
% The human evaluation table is refined as Tab.\ref{Tab:user}.
% \noindent\textbf{User Study.}
% We further randomly pick 3 videos from 10 videos generated by each model, 12 in total to be used in our user study. Every video has 3 dimensions of evaluation rating: Appearance, Consistency, Control, and the scores 1 to 5 indicate from low to high appearance, consistency and control capabilities. We recruit 30 users to evaluate each dimension of the 12 videos.
% The results are shown in Tab.\ref{Tab:user}. 
% The voters prefer our methods from all aspects compared with baselines. 
% Also, users prefer our method over Trailblazer in all dimensions.
% Moreover, when compared to MotionCtrl, users demonstrated a clear preference for the control capabilities of Motion-Zero. 
% Note that our proposed approach does not need any training. Thus, the baseline model's inherent limitations resulted in our approach being slightly inferior to MotionCtrl in aspects of consistency.

\begin{table}[t]
    \centering
    
    \resizebox{0.48\textwidth}{!}{%
    \begin{tabular}{cccc|cccc}
        \toprule
                  & Align & Cons. & Pick. & mIoU & AP50 & Cov. & CD($\downarrow$) \\
        \midrule
        w/o INPM   & 20.30          & 0.93      & 19.29 & 0.40 & 0.41 & 0.86 & 0.09 \\
        w/o SC & 21.27         & 0.92      & 19.10 & 0.18 & 0.17 & 0.71 & 0.22 \\
        w/o STAM  & 20.64          & 0.93     & 19.27 & 0.43 & 0.47 & 0.83 & 0.13 \\
        Ours & \textbf{21.96}          & \textbf{0.94}      & \textbf{19.89} & \textbf{0.54} & \textbf{0.64} & \textbf{0.96} & \textbf{0.07} \\
        \bottomrule
    \end{tabular}
    }
    \caption{The table of ablation study on different modules.}
    \vspace{-4mm}
    \label{tab:abl}
\end{table}

\subsection{Ablation Study}
\label{main:abl}
\noindent\textbf{Quantitative Analysis.}
We conduct ablation experiments on different components of Motion-Zero, as shown in Tab.\ref{tab:abl}. It indicates that removing individual modules leads to a significant decrease in the model's performance. 
We observe that when SC is removed, the scores of Align do not decrease significantly. However, when STAM or INPM is removed, the scores of Align decrease noticeably, demonstrating the significant role of the STAM and INPM modules in improving the consistency between text and generated content.
The SC has the greatest impact on location control as the scores drop from 0.54 to 0.18 in mIoU and from 0.64 to 0.17 in AP50.
Without the SC module, there is no guarantee for objects' position and spatial consistency between frames.
Additionally, the absence of STAM and INPM also results in inferior performance in temporal consistency for moving objects as the Cov. drops and it becomes difficult for the model to generate expected positioned objects as mIoU and AP50 decrease. Overall, the three modules are all important for video generation and the STAM and INPM have more impact on generation quality.

\noindent\textbf{Visualization of Attention Maps.}
% \section{Visualization of the attention maps}
\begin{figure}[tbp]
  \centering
  \includegraphics[width=0.8\linewidth]{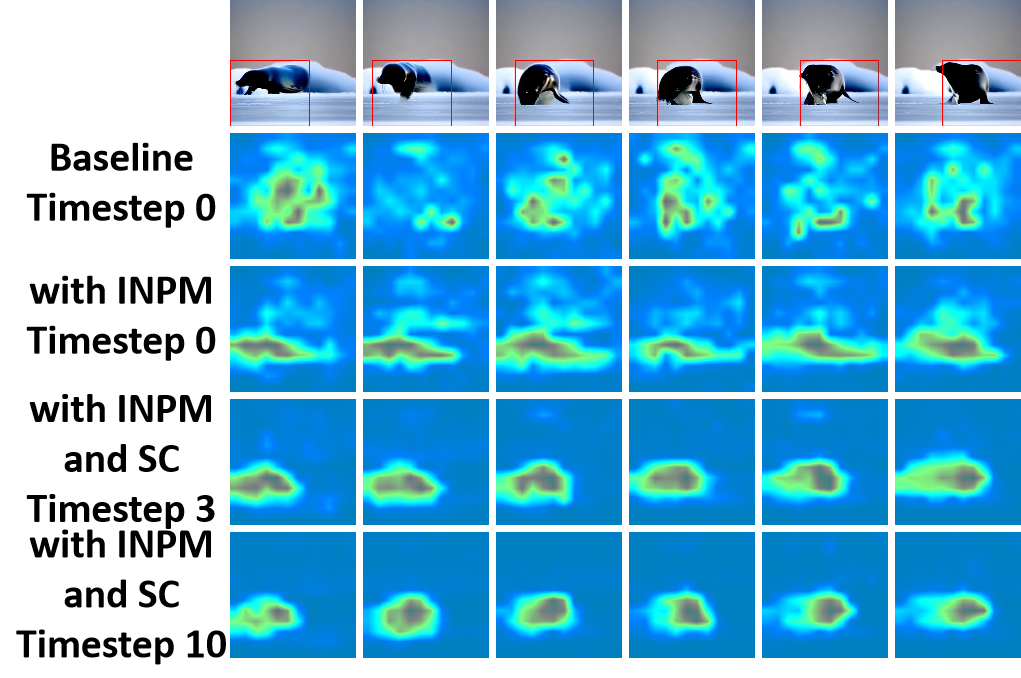}
      \captionof{figure}{Attention maps with different components. Prompt: \textit{A seal walking on the ice.}}
      \label{fig:attn}
\vspace{-5mm}
\end{figure}
We visualize the attention maps in the cross-attention block, as shown in Fig.\ref{fig:attn}. The second row displays the cross-attention map at step 0 under random initial noise conditions. It is observed that the attention values are random. As shown in the third row, when we introduce INPM, we can see that the areas with high attention response at step 0 are already approximately located at the positions of the user's box. This reflects that INPM can provide strong prior information for the position of moving objects. After optimization by SC, in the fourth and fifth rows, we can see that the areas with high response in the attention map are accurately located at the box position and are close to the position of the generated moving object. This demonstrates that SC can enforce high attention values inside the box while keeping the low response values outside the box.

\subsection{User Study}
\begin{table}[t]
    \centering
    \footnotesize
    
    \scalebox{0.95}{
    \begin{tabular}{cccc}
        \toprule
                  & Appearance & Consistency & Control \\
        \midrule
        
        ZeroScope & 2.97         & 2.80       &2.27  \\
        ModelScope & 2.57         & 2.55       &2.23  \\
        Trailblazer & 3.32         & 3.32        &3.40\\
        Peekaboo & 3.96         & 3.84        &3.55\\
        Ours & \textbf{4.75}      & \textbf{4.57}     &\textbf{4.67} \\
        
        \bottomrule
    \end{tabular}
    }
    \caption{The table of user study.}
    \label{Tab:user}
    \vspace{-5mm}
\end{table}

We randomly pick 3 videos from 10 videos generated by each model, 15 in total to be used in our user study. Every video has 3 dimensions of evaluation rating: Appearance, Consistency, and Control, and the scores 1 to 5 indicate appearance, consistency, and control capabilities from low to high. We randomly choose 60 users from our university through an incentivized questionnaire to evaluate each dimension of the 15 videos.
At the beginning of our questionnaire, we provide users with a detailed explanation of the task and the scoring guidelines. Only after confirming that the user fully understands these instructions do they proceed with the questionnaire.
The results are shown in Tab.\ref{Tab:user}. 
For baseline models, additional prompts are used to control the movements.
The voters prefer our methods from all aspects compared with baselines, Trailblazer, and Peekaboo.
To further validate the effectiveness of our user study, we use the Cronbach’s $\alpha$ coefficient to assess the internal consistency (reliability) of the questionnaire. The Cronbach’s $\alpha$ coefficient for our questionnaire is 0.901, indicating a high level of reliability. We use the Friedman Test to confirm that our results are statistically significant. The results indicate significant differences among the five methods (Friedman Test, $X^2$ = 1534.29, $p$ < 0.001).

\section{Conclusion}
In this paper, we proposed a novel zero-shot framework, Motion-Zero, for arbitrary object motion trajectory control that can be applied to various video diffusion models. 
Unlike previous methods require extensive training, our method enabled motion-control video generation without any fine-tuning of the baseline model. 
Our Initial Noise Prior Module was proposed to acquire prior initial noise for a high-quality generation. Moreover, Spatial Constraints and Shift Temporal Attention Mechanism respectively exploited the cross-attention dimension and temporal-attention dimension to obtain spatial control and temporal consistency.
Extensive experiments demonstrated the efficacy and generalization of our proposed approach.

\section{Acknowledgments}
This work is supported by the National Natural Science Foundation of China (62472178, 62376244), Fundamental Research Funds for the Central Universities, and Shanghai Urban Digital Transformation Special Fund Project (202301027). This work is also sponsored by Natural Science Foundation of Chongqing, China (CSTB2022NSCQ-MSX0552) and the Open Projects Program of State Key Laboratory of Multimodal Artificial Intelligence Systems (No.MAIS2024111).

\bibliography{aaai25}

\begin{thebibliography}{52}
\providecommand{\natexlab}[1]{#1}

\bibitem[{Bain et~al.(2021)Bain, Nagrani, Varol, and Zisserman}]{Bain21}
Bain, M.; Nagrani, A.; Varol, G.; and Zisserman, A. 2021.
\newblock Frozen in Time: A Joint Video and Image Encoder for End-to-End Retrieval.
\newblock In \emph{IEEE International Conference on Computer Vision}.

\bibitem[{Blattmann et~al.(2023)Blattmann, Dockhorn, Kulal, Mendelevitch, Kilian, Lorenz, Levi, English, Voleti, Letts et~al.}]{blattmann2023stable}
Blattmann, A.; Dockhorn, T.; Kulal, S.; Mendelevitch, D.; Kilian, M.; Lorenz, D.; Levi, Y.; English, Z.; Voleti, V.; Letts, A.; et~al. 2023.
\newblock Stable video diffusion: Scaling latent video diffusion models to large datasets.
\newblock \emph{arXiv preprint arXiv:2311.15127}.

\bibitem[{Cai et~al.(2023)Cai, Chen, Guan, Lin, Lu, Wang, and He}]{cai2023explicit}
Cai, Y.; Chen, L.; Guan, H.; Lin, S.; Lu, C.; Wang, C.; and He, G. 2023.
\newblock Explicit invariant feature induced cross-domain crowd counting.
\newblock In \emph{Proceedings of the AAAI Conference on Artificial Intelligence}, volume~37, 259--267.

\bibitem[{Cai et~al.(2021)Cai, Chen, Ma, Lu, Wang, and He}]{cai2021leveraging}
Cai, Y.; Chen, L.; Ma, Z.; Lu, C.; Wang, C.; and He, G. 2021.
\newblock Leveraging intra-domain knowledge to strengthen cross-domain crowd counting.
\newblock In \emph{2021 IEEE International Conference on Multimedia and Expo (ICME)}, 1--6. IEEE.

\bibitem[{Cai et~al.(2022)Cai, Ma, Lu, Wang, and He}]{cai2022global}
Cai, Y.; Ma, Z.; Lu, C.; Wang, C.; and He, G. 2022.
\newblock Global Representation Guided Adaptive Fusion Network for Stable Video Crowd Counting.
\newblock \emph{IEEE Transactions on Multimedia}, 25: 5222--5233.

\bibitem[{Chen et~al.(2023{\natexlab{a}})Chen, Cai, Lu, Wang, and He}]{chen2023video}
Chen, L.; Cai, Y.; Lu, C.; Wang, C.; and He, G. 2023{\natexlab{a}}.
\newblock Video-based spatio-temporal scene graph generation with efficient self-supervision tasks.
\newblock \emph{Multimedia Tools and Applications}, 82(25): 38947--38966.

\bibitem[{Chen et~al.(2024)Chen, Song, Cai, Lu, Li, Xie, Wang, and He}]{chen2024multi}
Chen, L.; Song, Y.; Cai, Y.; Lu, J.; Li, Y.; Xie, Y.; Wang, C.; and He, G. 2024.
\newblock Multi-Prototype Space Learning for Commonsense-Based Scene Graph Generation.
\newblock In \emph{Proceedings of the AAAI Conference on Artificial Intelligence}, volume~38, 1129--1137.

\bibitem[{Chen et~al.(2023{\natexlab{b}})Chen, Li, Wu, Jiang, Ding, and Zhang}]{NEURIPS2023_a24cd16b}
Chen, Z.; Li, B.; Wu, S.; Jiang, K.; Ding, S.; and Zhang, W. 2023{\natexlab{b}}.
\newblock Content-based Unrestricted Adversarial Attack.
\newblock In Oh, A.; Naumann, T.; Globerson, A.; Saenko, K.; Hardt, M.; and Levine, S., eds., \emph{Advances in Neural Information Processing Systems}, volume~36, 51719--51733. Curran Associates, Inc.

\bibitem[{Dhariwal and Nichol(2021)}]{dhariwal2021diffusion}
Dhariwal, P.; and Nichol, A. 2021.
\newblock Diffusion models beat gans on image synthesis.
\newblock \emph{Advances in neural information processing systems}, 34: 8780--8794.

\bibitem[{Epstein et~al.(2023)Epstein, Jabri, Poole, Efros, and Holynski}]{epstein2023diffusion}
Epstein, D.; Jabri, A.; Poole, B.; Efros, A.~A.; and Holynski, A. 2023.
\newblock Diffusion self-guidance for controllable image generation.
\newblock \emph{arXiv preprint arXiv:2306.00986}.

\bibitem[{Guo et~al.(2023)Guo, Yang, Rao, Wang, Qiao, Lin, and Dai}]{guo2023animatediff}
Guo, Y.; Yang, C.; Rao, A.; Wang, Y.; Qiao, Y.; Lin, D.; and Dai, B. 2023.
\newblock Animatediff: Animate your personalized text-to-image diffusion models without specific tuning.
\newblock \emph{arXiv preprint arXiv:2307.04725}.

\bibitem[{Hessel et~al.(2021)Hessel, Holtzman, Forbes, Bras, and Choi}]{hessel2021clipscore}
Hessel, J.; Holtzman, A.; Forbes, M.; Bras, R.~L.; and Choi, Y. 2021.
\newblock Clipscore: A reference-free evaluation metric for image captioning.
\newblock \emph{arXiv preprint arXiv:2104.08718}.

\bibitem[{Ho et~al.(2022)Ho, Chan, Saharia, Whang, Gao, Gritsenko, Kingma, Poole, Norouzi, Fleet et~al.}]{ho2022imagen}
Ho, J.; Chan, W.; Saharia, C.; Whang, J.; Gao, R.; Gritsenko, A.; Kingma, D.~P.; Poole, B.; Norouzi, M.; Fleet, D.~J.; et~al. 2022.
\newblock Imagen video: High definition video generation with diffusion models.
\newblock \emph{arXiv preprint arXiv:2210.02303}.

\bibitem[{Ho, Jain, and Abbeel(2020)}]{ho2020denoising}
Ho, J.; Jain, A.; and Abbeel, P. 2020.
\newblock Denoising diffusion probabilistic models.
\newblock \emph{Advances in neural information processing systems}, 33: 6840--6851.

\bibitem[{Jain et~al.(2024)Jain, Nasery, Vineet, and Behl}]{jain2024peekaboo}
Jain, Y.; Nasery, A.; Vineet, V.; and Behl, H. 2024.
\newblock Peekaboo: Interactive video generation via masked-diffusion.
\newblock In \emph{Proceedings of the IEEE/CVF Conference on Computer Vision and Pattern Recognition}, 8079--8088.

\bibitem[{Kirstain et~al.(2023)Kirstain, Polyak, Singer, Matiana, Penna, and Levy}]{kirstain2023pick}
Kirstain, Y.; Polyak, A.; Singer, U.; Matiana, S.; Penna, J.; and Levy, O. 2023.
\newblock Pick-a-pic: An open dataset of user preferences for text-to-image generation.
\newblock \emph{arXiv preprint arXiv:2305.01569}.

\bibitem[{Li et~al.(2024{\natexlab{a}})Li, Wang, Cai, Qi, Wang, Zhang, Song, Jiang, Huang, and Wang}]{li2024unifiedmllm}
Li, Z.; Wang, W.; Cai, Y.; Qi, X.; Wang, P.; Zhang, D.; Song, H.; Jiang, B.; Huang, Z.; and Wang, T. 2024{\natexlab{a}}.
\newblock Unifiedmllm: Enabling unified representation for multi-modal multi-tasks with large language model.
\newblock \emph{arXiv preprint arXiv:2408.02503}.

\bibitem[{Li et~al.(2024{\natexlab{b}})Li, Xu, Zhang, Song, Cai, Qi, Zhou, Pan, Li, Tu et~al.}]{li2024groundinggpt}
Li, Z.; Xu, Q.; Zhang, D.; Song, H.; Cai, Y.; Qi, Q.; Zhou, R.; Pan, J.; Li, Z.; Tu, V.; et~al. 2024{\natexlab{b}}.
\newblock Groundinggpt: Language enhanced multi-modal grounding model.
\newblock In \emph{Proceedings of the 62nd Annual Meeting of the Association for Computational Linguistics (Volume 1: Long Papers)}, 6657--6678.

\bibitem[{Ma, Lewis, and Kleijn(2023)}]{Ma2023TrailBlazerTC}
Ma, W.-D.~K.; Lewis, J.~P.; and Kleijn, W. 2023.
\newblock TrailBlazer: Trajectory Control for Diffusion-Based Video Generation.

\bibitem[{Minderer et~al.(2022)Minderer, Gritsenko, Stone, Neumann, Weissenborn, Dosovitskiy, Mahendran, Arnab, Dehghani, Shen, Wang, Zhai, Kipf, and Houlsby}]{minderer2022simple}
Minderer, M.; Gritsenko, A.; Stone, A.; Neumann, M.; Weissenborn, D.; Dosovitskiy, A.; Mahendran, A.; Arnab, A.; Dehghani, M.; Shen, Z.; Wang, X.; Zhai, X.; Kipf, T.; and Houlsby, N. 2022.
\newblock Simple Open-Vocabulary Object Detection with Vision Transformers.
\newblock \emph{arXiv preprint arXiv:2205.06230}.

\bibitem[{Qiu et~al.(2024)Qiu, Chen, Wang, He, Xia, and Liu}]{qiu2024freetraj}
Qiu, H.; Chen, Z.; Wang, Z.; He, Y.; Xia, M.; and Liu, Z. 2024.
\newblock FreeTraj: Tuning-Free Trajectory Control in Video Diffusion Models.
\newblock arXiv:2406.16863.

\bibitem[{Rombach et~al.(2022)Rombach, Blattmann, Lorenz, Esser, and Ommer}]{rombach2022high}
Rombach, R.; Blattmann, A.; Lorenz, D.; Esser, P.; and Ommer, B. 2022.
\newblock High-resolution image synthesis with latent diffusion models.
\newblock In \emph{Proceedings of the IEEE/CVF conference on computer vision and pattern recognition}, 10684--10695.

\bibitem[{Shi et~al.(2024)Shi, Huang, Wang, Bian, Li, Zhang, Zhang, Cheung, See, Qin et~al.}]{shi2024motion}
Shi, X.; Huang, Z.; Wang, F.-Y.; Bian, W.; Li, D.; Zhang, Y.; Zhang, M.; Cheung, K.~C.; See, S.; Qin, H.; et~al. 2024.
\newblock Motion-i2v: Consistent and controllable image-to-video generation with explicit motion modeling.
\newblock In \emph{ACM SIGGRAPH 2024 Conference Papers}, 1--11.

\bibitem[{Song, Meng, and Ermon(2020)}]{song2020denoising}
Song, J.; Meng, C.; and Ermon, S. 2020.
\newblock Denoising diffusion implicit models.
\newblock \emph{arXiv preprint arXiv:2010.02502}.

\bibitem[{Song et~al.(2023{\natexlab{a}})Song, Chen, Li, and Wang}]{10.1145/3544548.3581067}
Song, S.; Chen, J.; Li, C.; and Wang, C. 2023{\natexlab{a}}.
\newblock GVQA: Learning to Answer Questions about Graphs with Visualizations via Knowledge Base.
\newblock In \emph{Proceedings of the 2023 CHI Conference on Human Factors in Computing Systems}, CHI '23. New York, NY, USA: Association for Computing Machinery.
\newblock ISBN 9781450394215.

\bibitem[{Song et~al.(2024)Song, Li, Li, Chen, and Wang}]{9966829}
Song, S.; Li, C.; Li, D.; Chen, J.; and Wang, C. 2024.
\newblock GraphDecoder: Recovering Diverse Network Graphs From Visualization Images via Attention-Aware Learning.
\newblock \emph{IEEE Transactions on Visualization and Computer Graphics}, 30(7): 3074--3088.

\bibitem[{Song et~al.(2023{\natexlab{b}})Song, Li, Sun, and Wang}]{9720180}
Song, S.; Li, C.; Sun, Y.; and Wang, C. 2023{\natexlab{b}}.
\newblock VividGraph: Learning to Extract and Redesign Network Graphs From Visualization Images.
\newblock \emph{IEEE Transactions on Visualization and Computer Graphics}, 29(7): 3169--3181.

\bibitem[{Sterling(2023)}]{Sterling2023ZeroScope}
Sterling, S. 2023.
\newblock {ZeroScope}.

\bibitem[{Sun, Li, and Wang(2023)}]{10094642}
Sun, Y.; Li, Y.; and Wang, C. 2023.
\newblock Multi-Source Templates Learning for Real-Time Aerial Tracking.
\newblock In \emph{ICASSP 2023 - 2023 IEEE International Conference on Acoustics, Speech and Signal Processing (ICASSP)}, 1--5.

\bibitem[{Sun et~al.(2024{\natexlab{a}})Sun, Yu, Chen, Zhang, Huang, Li, Li, and Wang}]{sun2024chattracker}
Sun, Y.; Yu, F.; Chen, S.; Zhang, Y.; Huang, J.; Li, C.; Li, Y.; and Wang, C. 2024{\natexlab{a}}.
\newblock Chattracker: Enhancing visual tracking performance via chatting with multimodal large language model.
\newblock \emph{arXiv preprint arXiv:2411.01756}.

\bibitem[{Sun et~al.(2024{\natexlab{b}})Sun, Chen, Yao, Yi, Ding, and Ma}]{sun2024rethinking}
Sun, Z.; Chen, S.; Yao, T.; Yi, R.; Ding, S.; and Ma, L. 2024{\natexlab{b}}.
\newblock Rethinking Open-World DeepFake Attribution with Multi-perspective Sensory Learning.
\newblock \emph{International Journal of Computer Vision}.

\bibitem[{Sun et~al.(2023)Sun, Chen, Yao, Yin, Yi, Ding, and Ma}]{sun2023contrastive}
Sun, Z.; Chen, S.; Yao, T.; Yin, B.; Yi, R.; Ding, S.; and Ma, L. 2023.
\newblock Contrastive Pseudo Learning for Open-World DeepFake Attribution.
\newblock In \emph{Proceedings of the IEEE/CVF International Conference on Computer Vision (ICCV)}, 20882--20892.

\bibitem[{Wang et~al.(2023{\natexlab{a}})Wang, Li, Wu, Shen, Liu, Ding, and Zhou}]{wang2023rethinking}
Wang, H.; Li, B.; Wu, S.; Shen, S.; Liu, F.; Ding, S.; and Zhou, A. 2023{\natexlab{a}}.
\newblock Rethinking the learning paradigm for dynamic facial expression recognition.
\newblock In \emph{Proceedings of the IEEE/CVF conference on computer vision and pattern recognition}, 17958--17968.

\bibitem[{Wang et~al.(2023{\natexlab{b}})Wang, Yuan, Chen, Zhang, Wang, and Zhang}]{wang2023modelscope}
Wang, J.; Yuan, H.; Chen, D.; Zhang, Y.; Wang, X.; and Zhang, S. 2023{\natexlab{b}}.
\newblock Modelscope text-to-video technical report.
\newblock \emph{arXiv preprint arXiv:2308.06571}.

\bibitem[{Wang et~al.(2024{\natexlab{a}})Wang, Zhang, Zou, Zeng, Wei, Yuan, and Li}]{wang2024boximator}
Wang, J.; Zhang, Y.; Zou, J.; Zeng, Y.; Wei, G.; Yuan, L.; and Li, H. 2024{\natexlab{a}}.
\newblock Boximator: Generating rich and controllable motions for video synthesis.
\newblock \emph{arXiv preprint arXiv:2402.01566}.

\bibitem[{Wang et~al.(2024{\natexlab{b}})Wang, Li, Xu, Cai, Song, Qi, Zhou, Huang, Wang, and Xiao}]{wang2024qcrd}
Wang, W.; Li, Z.; Xu, Q.; Cai, Y.; Song, H.; Qi, Q.; Zhou, R.; Huang, Z.; Wang, T.; and Xiao, L. 2024{\natexlab{b}}.
\newblock QCRD: Quality-guided Contrastive Rationale Distillation for Large Language Models.
\newblock \emph{arXiv preprint arXiv:2405.13014}.

\bibitem[{Wang et~al.(2024{\natexlab{c}})Wang, Li, Xu, Li, Cai, Jiang, Song, Hu, Wang, and Xiao}]{wang2024advancing}
Wang, W.; Li, Z.; Xu, Q.; Li, L.; Cai, Y.; Jiang, B.; Song, H.; Hu, X.; Wang, P.; and Xiao, L. 2024{\natexlab{c}}.
\newblock Advancing Fine-Grained Visual Understanding with Multi-Scale Alignment in Multi-Modal Models.
\newblock \emph{arXiv preprint arXiv:2411.09691}.

\bibitem[{Wang et~al.(2023{\natexlab{c}})Wang, Yuan, Zhang, Chen, Wang, Zhang, Shen, Zhao, and Zhou}]{wang2023videocomposer}
Wang, X.; Yuan, H.; Zhang, S.; Chen, D.; Wang, J.; Zhang, Y.; Shen, Y.; Zhao, D.; and Zhou, J. 2023{\natexlab{c}}.
\newblock VideoComposer: Compositional Video Synthesis with Motion Controllability.
\newblock \emph{arXiv preprint arXiv:2306.02018}.

\bibitem[{Wang et~al.(2023{\natexlab{d}})Wang, Yuan, Wang, Chen, Xia, Luo, and Shan}]{wang2023motionctrl}
Wang, Z.; Yuan, Z.; Wang, X.; Chen, T.; Xia, M.; Luo, P.; and Shan, Y. 2023{\natexlab{d}}.
\newblock MotionCtrl: A Unified and Flexible Motion Controller for Video Generation.
\newblock \emph{arXiv preprint arXiv:2312.03641}.

\bibitem[{Wu et~al.(2024)Wu, Li, Zeng, Zhang, Zhou, Li, Tong, and Chen}]{wu2024motionbooth}
Wu, J.; Li, X.; Zeng, Y.; Zhang, J.; Zhou, Q.; Li, Y.; Tong, Y.; and Chen, K. 2024.
\newblock Motionbooth: Motion-aware customized text-to-video generation.
\newblock \emph{arXiv preprint arXiv:2406.17758}.

\bibitem[{Wu et~al.(2023{\natexlab{a}})Wu, Ge, Wang, Lei, Gu, Shi, Hsu, Shan, Qie, and Shou}]{wu2023tune}
Wu, J.~Z.; Ge, Y.; Wang, X.; Lei, S.~W.; Gu, Y.; Shi, Y.; Hsu, W.; Shan, Y.; Qie, X.; and Shou, M.~Z. 2023{\natexlab{a}}.
\newblock Tune-a-video: One-shot tuning of image diffusion models for text-to-video generation.
\newblock In \emph{Proceedings of the IEEE/CVF International Conference on Computer Vision}, 7623--7633.

\bibitem[{Wu et~al.(2023{\natexlab{b}})Wu, Li, Gao, Dong, Bai, Singh, Xiang, Li, Huang, Sun, He, Hu, Hu, Huang, Zhu, Cheng, Tang, Shou, Keutzer, and Iandola}]{wu2023cvpr}
Wu, J.~Z.; Li, X.; Gao, D.; Dong, Z.; Bai, J.; Singh, A.; Xiang, X.; Li, Y.; Huang, Z.; Sun, Y.; He, R.; Hu, F.; Hu, J.; Huang, H.; Zhu, H.; Cheng, X.; Tang, J.; Shou, M.~Z.; Keutzer, K.; and Iandola, F. 2023{\natexlab{b}}.
\newblock CVPR 2023 Text Guided Video Editing Competition.
\newblock arXiv:2310.16003.

\bibitem[{Xie et~al.(2023)Xie, Li, Huang, Liu, Zhang, Zheng, and Shou}]{xie2023boxdiff}
Xie, J.; Li, Y.; Huang, Y.; Liu, H.; Zhang, W.; Zheng, Y.; and Shou, M.~Z. 2023.
\newblock Boxdiff: Text-to-image synthesis with training-free box-constrained diffusion.
\newblock In \emph{Proceedings of the IEEE/CVF International Conference on Computer Vision}, 7452--7461.

\bibitem[{Xue et~al.(2022)Xue, Hang, Zeng, Sun, Liu, Yang, Fu, and Guo}]{xue2022advancing}
Xue, H.; Hang, T.; Zeng, Y.; Sun, Y.; Liu, B.; Yang, H.; Fu, J.; and Guo, B. 2022.
\newblock Advancing high-resolution video-language representation with large-scale video transcriptions.
\newblock In \emph{Proceedings of the IEEE/CVF Conference on Computer Vision and Pattern Recognition}, 5036--5045.

\bibitem[{Yang, Li, and Chen(2024)}]{ijcai2024p762}
Yang, L.; Li, Y.; and Chen, L. 2024.
\newblock ClothPPO: A Proximal Policy Optimization Enhancing Framework for Robotic Cloth Manipulation with Observation-Aligned Action Spaces.
\newblock In Larson, K., ed., \emph{Proceedings of the Thirty-Third International Joint Conference on Artificial Intelligence, {IJCAI-24}}, 6895--6903. International Joint Conferences on Artificial Intelligence Organization.
\newblock Main Track.

\bibitem[{Yin et~al.(2023)Yin, Wu, Liang, Shi, Li, Ming, and Duan}]{yin2023dragnuwa}
Yin, S.; Wu, C.; Liang, J.; Shi, J.; Li, H.; Ming, G.; and Duan, N. 2023.
\newblock Dragnuwa: Fine-grained control in video generation by integrating text, image, and trajectory.
\newblock \emph{arXiv preprint arXiv:2308.08089}.

\bibitem[{Yu et~al.(2024)Yu, Fang, Zheng, Sigurdsson, Ordonez, Piramuthu, and Bansal}]{yu2024zero}
Yu, S.; Fang, J.~Z.; Zheng, J.; Sigurdsson, G.; Ordonez, V.; Piramuthu, R.; and Bansal, M. 2024.
\newblock Zero-shot controllable image-to-video animation via motion decomposition.
\newblock In \emph{Proceedings of the 32nd ACM International Conference on Multimedia}, 3332--3341.

\bibitem[{Zhang et~al.(2023)Zhang, Wu, Liu, Zhao, Ran, Gu, Gao, and Shou}]{zhang2023show}
Zhang, D.~J.; Wu, J.~Z.; Liu, J.-W.; Zhao, R.; Ran, L.; Gu, Y.; Gao, D.; and Shou, M.~Z. 2023.
\newblock Show-1: Marrying pixel and latent diffusion models for text-to-video generation.
\newblock \emph{arXiv preprint arXiv:2309.15818}.

\bibitem[{Zhang et~al.(2017)Zhang, Cisse, Dauphin, and Lopez-Paz}]{zhang2017mixup}
Zhang, H.; Cisse, M.; Dauphin, Y.~N.; and Lopez-Paz, D. 2017.
\newblock mixup: Beyond empirical risk minimization.
\newblock \emph{arXiv preprint arXiv:1710.09412}.

\bibitem[{Zhang, Rao, and Agrawala(2023)}]{Zhang2023AddingCC}
Zhang, L.; Rao, A.; and Agrawala, M. 2023.
\newblock Adding Conditional Control to Text-to-Image Diffusion Models.
\newblock \emph{ArXiv}, abs/2302.05543.

\bibitem[{Zheng et~al.(2024{\natexlab{a}})Zheng, Geng, Jiang, Wan, Zhang, Chen, Wang, and Li}]{DBLP:conf/mm/Zheng-ddpmbs-mm}
Zheng, T.; Geng, C.; Jiang, P.; Wan, B.; Zhang, H.; Chen, J.; Wang, J.; and Li, B. 2024{\natexlab{a}}.
\newblock Non-uniform Timestep Sampling: Towards Faster Diffusion Model Training.
\newblock In Cai, J.; Kankanhalli, M.~S.; Prabhakaran, B.; Boll, S.; Subramanian, R.; Zheng, L.; Singh, V.~K.; C{\'{e}}sar, P.; Xie, L.; and Xu, D., eds., \emph{Proceedings of the 32nd {ACM} International Conference on Multimedia, {MM} 2024, Melbourne, VIC, Australia, 28 October 2024 - 1 November 2024}, 7036--7045. {ACM}.

\bibitem[{Zheng et~al.(2024{\natexlab{b}})Zheng, Jiang, Wan, Zhang, Chen, Wang, and Li}]{DBLP:conf/eccv/Zheng-bttdm-eccv}
Zheng, T.; Jiang, P.; Wan, B.; Zhang, H.; Chen, J.; Wang, J.; and Li, B. 2024{\natexlab{b}}.
\newblock Beta-Tuned Timestep Diffusion Model.
\newblock In Leonardis, A.; Ricci, E.; Roth, S.; Russakovsky, O.; Sattler, T.; and Varol, G., eds., \emph{Computer Vision - {ECCV} 2024 - 18th European Conference, Milan, Italy, September 29-October 4, 2024, Proceedings, Part {III}}, volume 15061 of \emph{Lecture Notes in Computer Science}, 114--130. Springer.

\end{thebibliography}

\newpage

\newpage

The appendix provides additional details and in-depth analyses of Motion-Zero. The code will be released after acceptance.
The structure of the appendix is as follows: 
\begin{itemize}
    \item Comparation with SOTA pre-trained methods.
    \item A more comprehensive ablation study. 
    \item The details of the evaluation dataset. 
    \item Discuss, limitations and future work. 
    \item Border Impacts. 
\end{itemize}

\section{Compare with Pre-trained SOTA methods}
\begin{figure*}[htbp]
  \centering
  \includegraphics[width=1\linewidth]{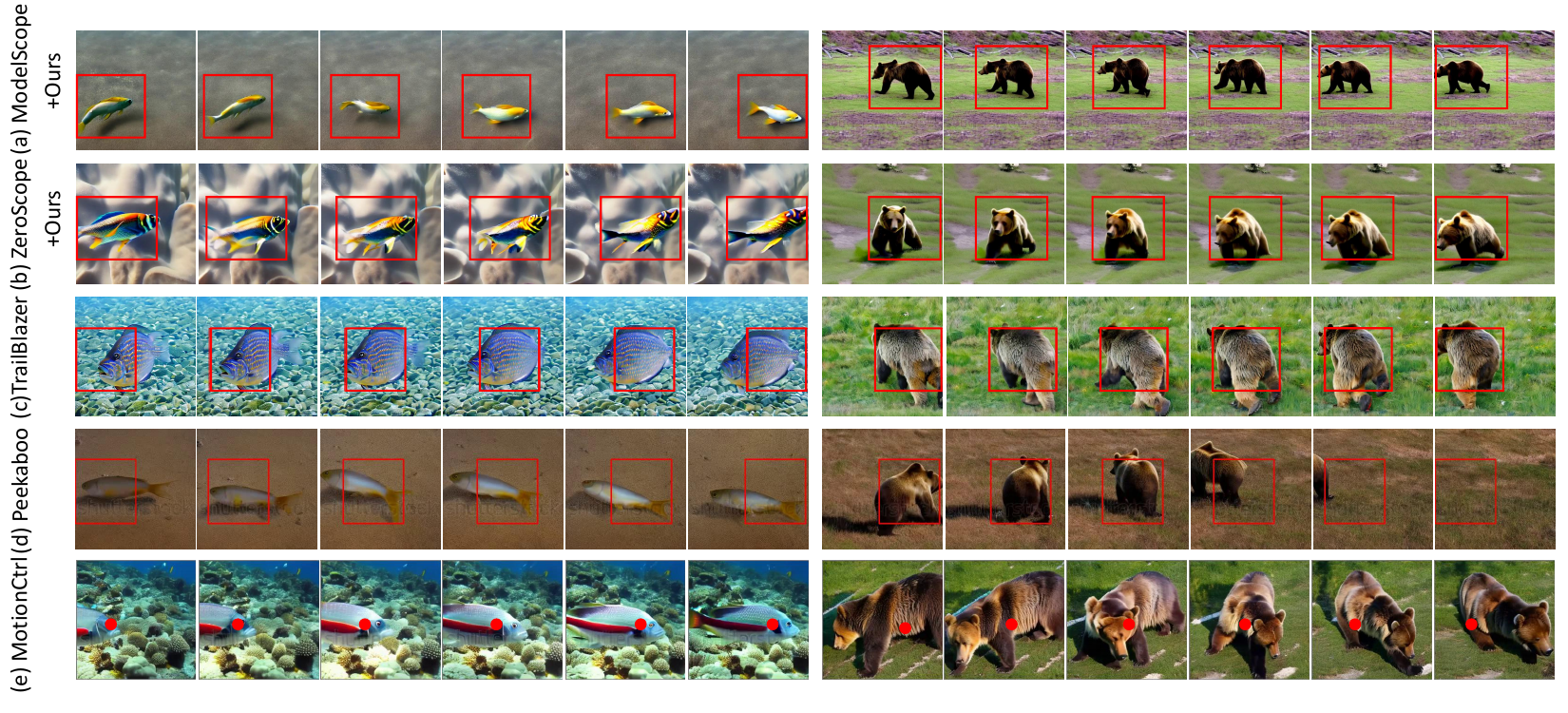}

   \caption{Comparisons with SOTA methods. The prompt of the left part is \textit{A fish is swimming in the sea.} Right: \textit{A bear is walking on the grass.} \textbf{Zoom in for the best view.}}
   \label{fig:supp_com}
%\vspace{-4mm}
\end{figure*}

To evaluate the effectiveness of our method, we compare it with the current state-of-the-art pre-trained method for controlling the trajectories of moving objects, MotionCtrl \cite{wang2023motionctrl}. As shown in Fig. \ref{fig:supp_com}, our results are comparable to those of the pre-trained model in terms of controlling the direction of the fish's movements. In the case of generating a bear, MotionCtrl only manages to rotate the bear's body, whereas Motion-Zero can effectively generate the bear's movement. Note that our method additionally allows for the control of their size. Our advantage also lies in the ability to plug and play with base models without the need for any training.

\section{Ablation Study}
\label{app:abl}

\subsubsection{Ablation on Main Components} 
We conduct a series of ablation experiments to verify the effectiveness of our modules individually.

\begin{figure}[htbp]
  \centering
  \includegraphics[width=1\linewidth]{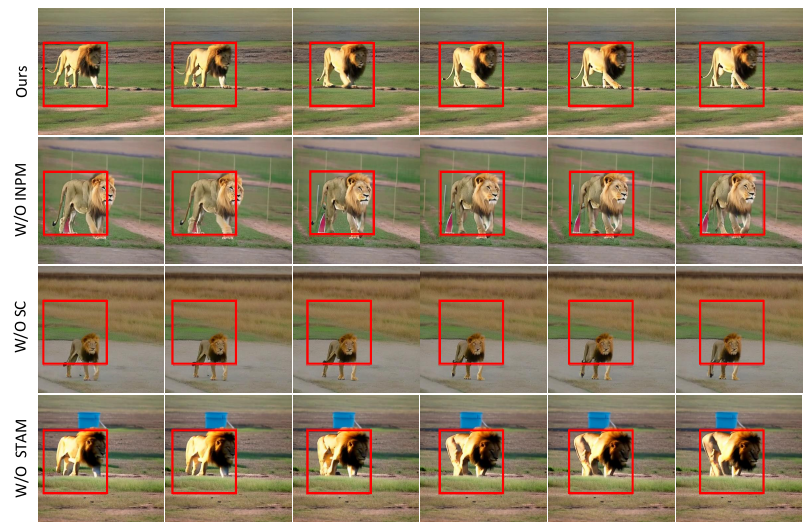}

   \caption{Ablation studies on different components. We use ZeroScope as our baseline model. To demonstrate the coherence of the generated object's motion, we captured every other frame, resulting in a total of 6 frames. The prompt is \textit{A lion is walking on the field.} \textbf{Zoom in for the best view.}}
   \label{fig:Ablation}
%\vspace{-4mm}
\end{figure}

\noindent\textbf{Impact of INPM.} INPM is designed to provide a prior for the initial position of moving objects, facilitating the model to more stably control the position of the object's movement. 
As shown in Fig.\ref{fig:Ablation}, the second row demonstrates the effects of removing this module. 
We can see that the lion is not entirely within the frame, and due to the forceful control by SC, the lion's body undergoes significant deformation, while the lion's head shows no movement at all. 
% In the right part, although the elephant is generated within the box, its appearance is greatly distorted, with even two elephants appearing.
For Motion-Zero, the lion's body is clearly generated, and the lion's head also moves with the changes in the frame, resulting in the high overall stability of the lion. 
This shows that the INPM module plays a significant role in stably controlling the position of objects and generating stable videos.

\noindent\textbf{Impact of SC.} 
In this section, we validate the importance of SC. SC primarily provides auxiliary support for control over spatial positioning. 
As shown in the third line in Fig.\ref{fig:Ablation}, without spatial constraints, the generated motion of the lion appears mostly outside the box. 
% In the right part of the same line, we can see that the centroid of the elephant tends to be on the left side rather than in the center of the box.
The SC is the most important module for our Motion-Zero controlling method.
The reason is that SC is the only gradient-based modification to the baseline compared with other modules.
% However, in the first row, the lion generated by the model is completely within the red box. 
% In the right part, the centroid of the elephant always close to the box center.
% These demonstrate that our SC module can achieve precise control over the position of object motion.

\noindent\textbf{Impact of STAM.}
The role of the STAM is to ensure the temporal consistency of the moving object, which is to maintain the continuity of its motion as it moves. 
As illustrated in Fig.\ref{fig:Ablation}, we present the effects of omitting the STAM module in the last row. 
It can be noted that during the movement, the lion's legs undergo deformation and become blurred. 
% The right part shows the elephant's leg movement becomes unstable, and in some frames, the elephant's legs are missing.
In contrast, the first row demonstrates that the lion's leg movements are well-generated with strong continuity, highlighting the significance of our STAM. More details are provided in the supplementary materials.

\subsubsection{Ablation on Parameter $T_1$}
\begin{figure}[htbp]
  \centering
  \includegraphics[width=1\linewidth]{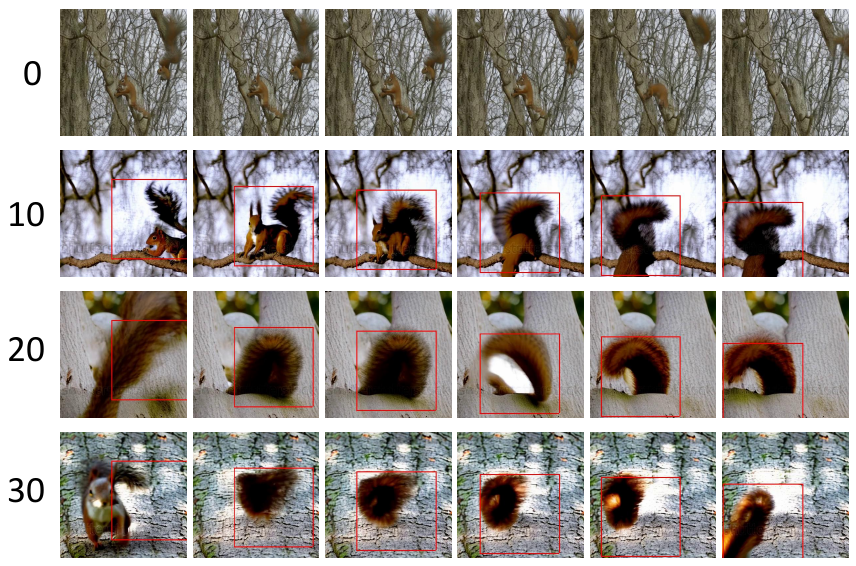}

   \caption{Ablation studies on $T_1$ ranging from 0 to 30. The prompt is \textit{A squirrel descending a tree after gathering nuts.} \textbf{Zoom in for the best view.}}
   \label{fig:T1}
%\vspace{-4mm}
\end{figure}

$T_1$ determines the timesteps at which the SC and STAM modules act during the diffusion denoising process. As shown in the fig. \ref{fig:T1}, when no constraints are applied, the squirrel essentially shows no displacement. When $T_1$ is set to 10, the squirrel moves with the box while maintaining its stability. When $T_1$ is 20, although the squirrel moves with the box, its shape does not remain consistent. At $T_1$ of 30, even though the moving object is still within the box, it is almost unrecognizable as a squirrel. The conclusion is that a higher $T_1$ provides better control but significantly reduces clarity and object identity maintenance. From the experiments, the optimal value for $T_1$ is 10.

\subsubsection{Ablation on Parameter $\lambda_i$ and $\lambda_o$}
\begin{figure}[htbp]
  \centering
   \includegraphics[width=1\linewidth]{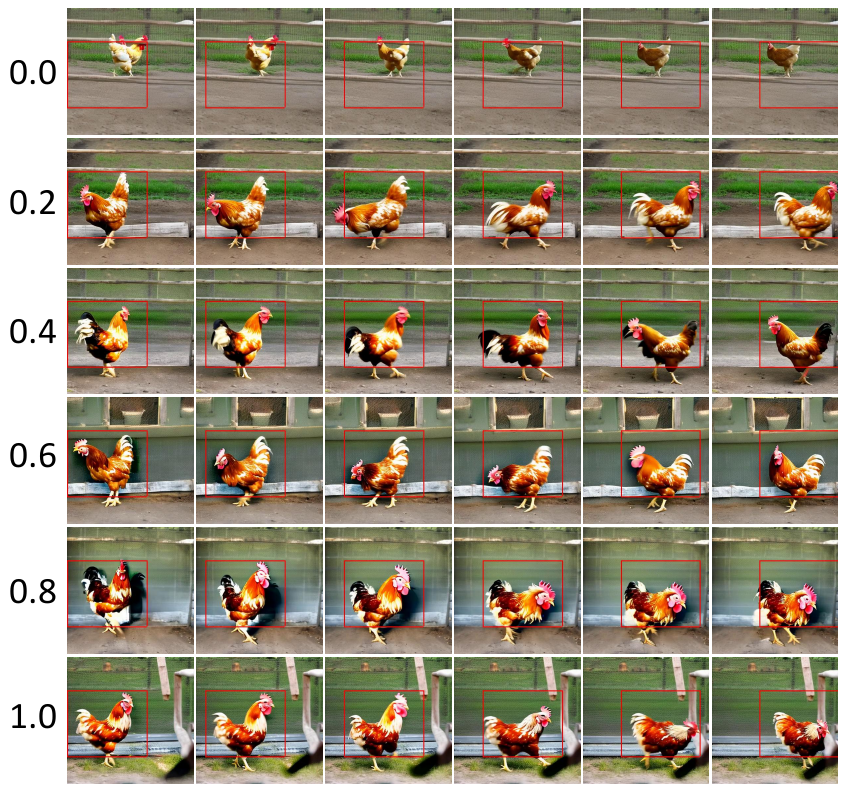}

   \caption{Ablation studies on different $\lambda_i$ and $\lambda_o$ range from 0.0 to 1.0. The prompt: \textit{A chicken is walking in the farm.}}
   \label{fig:box}
\end{figure}

In SC, $L_i$ and $L_o$ are designed to increase the values of $\mathbf{A}$ corresponding to the prompt inside the box and decrease the values of $\mathbf{A}$ for the prompt outside the box.
As depicted in Fig.\ref{fig:box}, the variations in the video generation effects with $\lambda_i$ and $\lambda_o$ values ranging from 0.0 to 1.0 were tested. 
From the first row, it can be seen that without the use of $\lambda_i$ and $\lambda_o$, the box does not effectively control the outcome.
As $\lambda_i$ and $\lambda_o$ are incremented, it is evident that the chicken increasingly appears concentrated within the box. In the final row, it is observed that the box can perfectly control the chicken's movement trajectory.

\subsubsection{Effectiveness of Parameter $\lambda_s$}
\begin{figure}[htbp]
  \centering
   \includegraphics[width=1\linewidth]{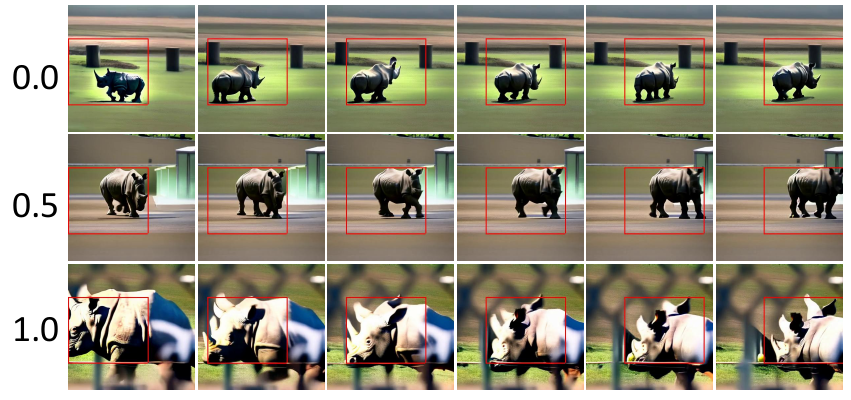}

   \caption{Ablation studies on different $\lambda_s$ values of 0.0, 0.5, 1.0. The prompt: \textit{A rhinoceros is walking on the field.}}
   \label{fig:sloss}
\end{figure}

In SC, $L_s$ is used to maintain the object's consistency across frames. 
As shown in the Fig.\ref{fig:sloss}, the effects of the output video were tested with $\lambda_s$ at 0.0, 0.5, and 1.0. 
It is observed that when $\lambda_s$ is set to 0.0, the orientation and some of the rhinoceros's external features do not maintain consistency. 
With $\lambda_s$ at 0.5, the appearance of the rhinoceros is observed to be consistent throughout. 
However, when $\lambda_s$ is increased to 1.0, it is noted that although the object within the box maintained consistency, the overall semantics of the video were disrupted. 
From the experiment, the optimal value for $\lambda_s$ was determined to be 0.5.

\subsubsection{Effectiveness of Parameter $\lambda_c$}
\begin{figure}[htbp]
  \centering
   \includegraphics[width=1\linewidth]{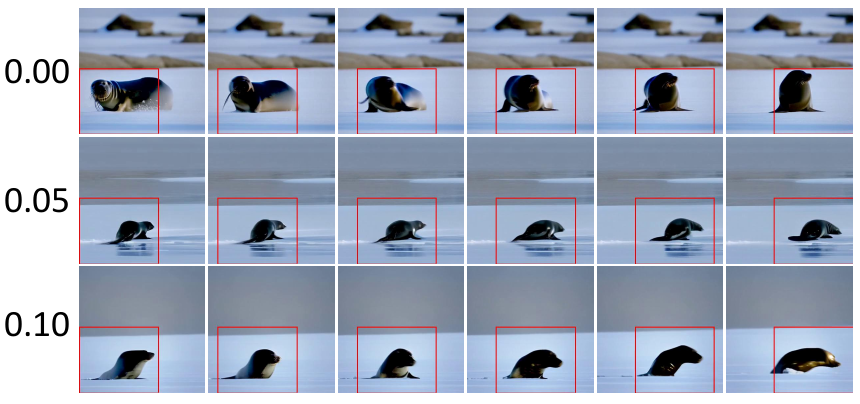}

   \caption{Ablation studies on different $\lambda_c$ values of 0.00, 0.05, 0.10. The prompt part: \textit{A seal is walking on the ice.}}
   \label{fig:closs}
\end{figure}

In SC, $L_c$ is utilized to ensure that the center of the object approaches the center of the given box. 
As illustrated in Fig.\ref{fig:closs}, evaluations are conducted to observe the control effects on video generation when $\lambda_c$ was set to 0.00, 0.05, and 0.10. 
It is demonstrated that with $\lambda_c$ at 0.00, although the seal remains inside the box, its center does not align with the center of the box, resulting in the seal not moving with the box but instead changing direction. 
At a $\lambda_c$ value of 0.05, the seal is observed to move perfectly with the motion of the box. 
When $\lambda_c$ is increased to 0.10, the background become blurry, and the seal's body appeared distorted. 
Based on the experimental evidence, a $\lambda_c$ value of 0.05 yielded the best control outcome.

\subsubsection{Influence of Parameter $\lambda_p$}
In the INPM, $\lambda$ is used to control the ratio between the DDIM Inversion \cite{song2020denoising} prior noise $\mathbf{z}_I$ and the standard normal noise $\mathbf{z}^*$. 
As shown in Fig.\ref{fig:prior}, we utilized two sets of prompts and varied the parameter $\lambda_p$ from 0.0 to 1.0 in increments of 0.2 to demonstrate the effects of different $\lambda_p$ values.
We observed that when the value of $\lambda_p$ is set to 0.0, the video featuring sheep on the left side shows multiple sheep appearing. 
On the right side, the pig does not appear within the frame. 
Instead of moving from left to right, it seems more like the pig is turning, with little change in the position of its head.
When $\lambda_p$ was increased to 0.2, it was noted that the sheep on the left side were reduced to a single individual, which represented an improvement. 
However, the positioning of the sheep did not effectively correspond with the movements of the frame. 
For the pig depicted on the right side, its position appeared to be relatively static, not following the shifts in the frame, with only the head showing movement through turning.
With $\lambda_p$ set between 0.4 and 1.0, the movement of the sheep on the left was found to be well-controlled. 
However, the motion trajectory of the pig on the right proved challenging to control at values of $\lambda_p$ at 0.4, 0.6, and 1.0. It was only when $\lambda_p$ reached 0.8 that the pig's movement trajectory was perfectly managed.
It can be concluded that the INPM plays a crucial role in controlling object positioning. 
Based on experimental observations, a $\lambda_p$ value of 0.8 has emerged as the optimal empirical setting.

\section{Details of Evaluation Datasets}
\label{app:eval}
In this section, we elaborate on the validation dataset described in Section 5.2.2 of the main text.

\noindent\textbf{Evaluation Prompts.} 
Similar to Trailblazer \cite{Ma2023TrailBlazerTC}, we used 33 prompts with various subjects and motions as the evaluation dataset, as shown below:
\begin{itemize}
    \item A woodpecker climbing up a tree trunk.
    \item A squirrel descending a tree after gathering nuts.
    \item A bird diving towards the water to catch fish.
    \item A frog leaping up to catch a fly.
    \item A parrot flying upwards towards the treetops.
    \item A squirrel jumping from one tree to another.
    \item A rabbit burrowing downwards into its warren.
    \item A satellite orbiting Earth in outer space.
    \item A skateboarder performing tricks at a skate park.
    \item A leaf falling gently from a tree.
    \item A paper plane gliding in the air.
    \item A bear climbing down a tree after spotting a threat.
    \item A duck diving underwater in search of food.
    \item A kangaroo is hopping down a gentle slope.
    \item An owl swooping down on its prey during the night.
    \item A balloon drifting across a clear sky.
    \item A bus moving through London streets.
    \item A plane flying high in the sky.
    \item A helicopter hovering above a cityscape.
    \item A streetcar trundling down tracks in a historic district.
    \item A rocket launching into space from a launchpad.
    \item A deer standing in a snowy field.
    \item A horse grazing in a meadow.
    \item A fox sitting in a forest clearing.
    \item A swan floating gracefully on a lake.
    \item A panda munching bamboo in a bamboo forest.
    \item A penguin standing on an iceberg.
    \item A lion lying in the savanna grass.
    \item An owl perched silently in a tree at night.
    \item A dolphin just breaking the ocean surface.
    \item A camel resting in a desert landscape.
    \item A kangaroo standing in the Australian outback.
    \item A colorful hot air balloon tethered to the ground.
\end{itemize}

\begin{figure}[htbp]
  \centering
   \includegraphics[width=1\linewidth]{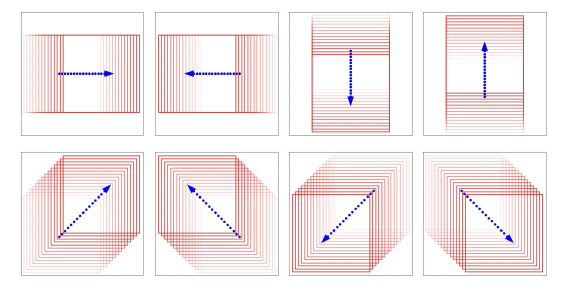}

   \caption{Simple trajectory evaluation dataset, consisting of eight simple trajectories. The blue points represent the center of the boxes. The color of the bounding box in the first frame is the lightest, while the color of the bounding box in the last frame is the darkest.}
   \label{fig:simple_traj}
\end{figure}

\begin{figure}[htbp]
  \centering
   \includegraphics[width=1\linewidth]{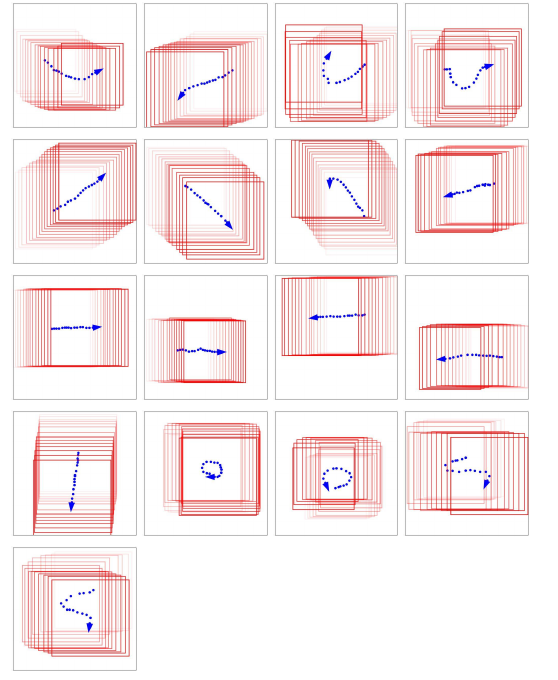}

   \caption{Complex trajectory evaluation dataset, consisting of seventeen complex trajectories. The blue points represent the center of the boxes. The color of the bounding box in the first frame is the lightest, while the color of the bounding box in the last frame is the darkest.}
   \label{fig:complex_traj}
\end{figure}

\begin{figure*}[htbp]
  \centering
   \includegraphics[width=1\linewidth]{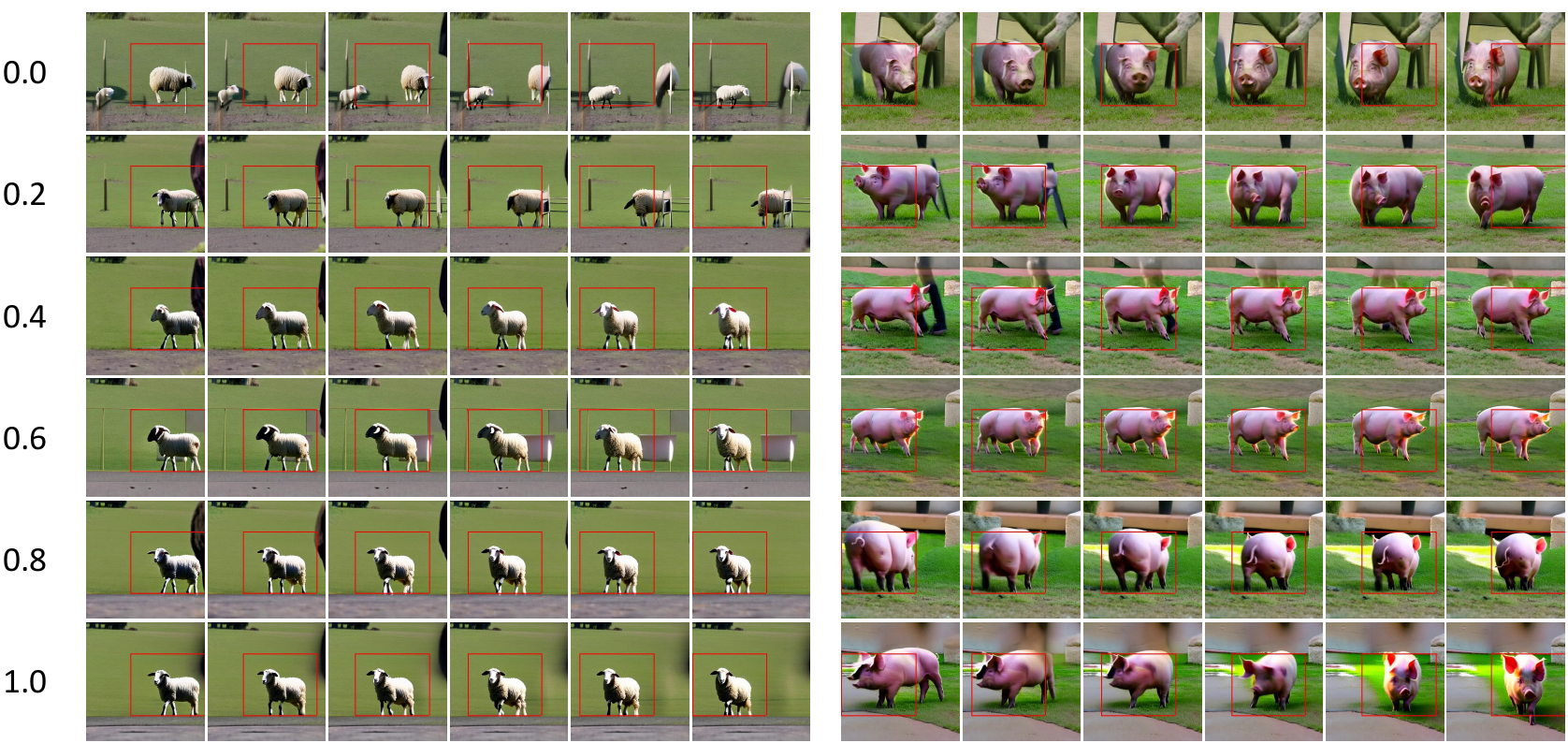}

   \caption{Ablation studies on different $\lambda_p$ range from 0.0 to 1.0. The prompt of the left part: \textit{A sheep is walking on the field.} Right part: \textit{A pig is walking on the grass.}}
   \label{fig:prior}
\end{figure*}

\noindent\textbf{Evaluation Trajectories.}
Following MotionCtrl \cite{wang2023motionctrl}, we use 8 simple trajectories (Fig.\ref{fig:simple_traj}) and 17 complex trajectories (Fig.\ref{fig:complex_traj}) to test the robustness of our method. 

\section{Discussion, Limitations and Future Works}
\label{lim}
Due to the absence of the need for additional training, our method can be applied to all pre-trained video diffusion models. 
However, on the other hand, its generative performance is entirely dependent on the base model. 
Existing video diffusion models sometimes generate relatively unstable objects, such as deformation in moving objects, holes in the background, and various issues like the inability to precisely control the attributes of generated objects with text.
% The hyper-parameters of our method require some fine-tuning for different base models.

The hyper-parameters presented in our paper are empirical values that we have found to be applicable in most cases. If we need to apply our algorithm to a new base model, appropriate parameter adjustments can yield better results.

Our method can currently effectively control the movement of objects in videos, but the trajectory of the movement is entirely controlled by the user, lacking semantic interaction with the video. 
For example, in a forest scene, if a user provides the first frame as a reference, they can use a prompt like \textit{a little rabbit goes to the stream to drink water} to control a rabbit to navigate and avoid obstacles in the forest automatically. 
This kind of interaction between object movement and the background, as well as narrative-like semantic control, will be the direction of our future work.

\section{Broader Impacts}
\label{BI}
Our approach is based on video diffusion models, and it does not require training, so in principle,  both potential positive societal impacts and negative societal impacts of our method on society are the same as the potential impacts that the video diffusion model may have on society. Furthermore, our approach can make the generation of controllable videos easier, allowing the generated videos to better match user expectations, and potentially improving the efficiency of social production.

\end{document}